\documentclass[journal]{IEEEtran}
\IEEEoverridecommandlockouts
\usepackage{cite}
\usepackage{amsmath,amssymb,amsfonts}
\usepackage{algorithmic}
\usepackage{graphicx}
\usepackage{textcomp}
\usepackage{xcolor}
\usepackage{algorithm}
\usepackage{bm}
\usepackage{subfigure}
\usepackage{multirow}
\usepackage{booktabs}
\usepackage{array}
\usepackage{makecell}
\usepackage{url} 
\usepackage{booktabs}
\usepackage[justification=centering]{caption}

\usepackage[defaultcolor=magenta]{changes}
\usepackage{subcaption}
\usepackage[hidelinks]{hyperref}
\usepackage{float} 

\def\BibTeX{{\rm B\kern-.05em{\sc i\kern-.025em b}\kern-.08em
    T\kern-.1667em\lower.7ex\hbox{E}\kern-.125emX}}
\begin{document}

\title{HAFLQ: Heterogeneous Adaptive Federated LoRA Fine-tuned LLM with Quantization}

\author{\IEEEauthorblockN{Yang~Su,\ Na~Yan,\ Yansha~Deng, Mischa Dohler, and Robert Schober}
	\thanks{Yang Su, Na Yan and Yansha Deng are with the Department of Engineering, King’s College London, London, WC2R 2LS, U.K. (email:\{yang.2.su, na.2.yan, yansha.deng\}@kcl.ac.uk). (Corresponding author: Yansha Deng). This paper will be presented in part at the 2025 IEEE International Conference on Communications, to appear\cite{su2024federated}.}
 \thanks{Mischa Dohler is with the Advanced Technology Group, Ericsson Inc., Silicon Valley, US. (email: mischa.dohler@ericsson.com).}
\thanks{Robert Schober is with the Institute for Digital Communication,
Friedrich-Alexander-Universit\"at Erlangen-N\"urnberg, 91052 Erlangen, Germany (e-mail: robert.schober@fau.de).}
}
\maketitle

\begin{abstract}
Federated fine-tuning of pre-trained Large Language Models (LLMs) enables task-specific adaptation across diverse datasets while preserving privacy. However, challenges such as high computational and memory demands, heterogeneous client resources, bandwidth constraints, and ineffective global aggregation hinder its efficiency. To address these issues, we propose HAFLQ (Heterogeneous Adaptive Federated Low-Rank Adaptation Fine-tuned LLM with Quantization), a novel framework for efficient and scalable federated fine-tuning of LLMs in heterogeneous environments.
To reduce memory and computation demands, we propose a salience-driven adaptive LLM quantization framework that evaluates the importance of transformer blocks using a salience metric and applies adaptive block-wise quantization accordingly.
To handle heterogeneous computational capabilities, we propose an importance-based parameter truncation and freezing scheme. 
To address communication bottlenecks, we propose an importance-aware bandwidth-adaptive quantization method, which dynamically adjusts parameter precision based on importance and bandwidth constraints.
To improve global model aggregation, we propose an adaptive rank-1 matrix-level aggregation strategy, which prevents information dilution and accelerates convergence by aggregating only updated rank-1 matrices from clients.
Experimental results on the text classification task demonstrate that HAFLQ reduces memory usage by 31\%, lowers communication cost by 49\%, improves accuracy by 50\%, and achieves faster convergence compared to the baseline method.

\end{abstract}

\begin{IEEEkeywords}
Large Language Model, Low-Rank Adaptation, Federated Learning, Quantization.
\end{IEEEkeywords}

\section{Introduction}
Large Language Models (LLMs) have exhibited exceptional performance in understanding and generating natural language across a wide range of tasks, including applications in chatbots and search engines\cite{zhao2023survey, chang2024survey,naveed2023comprehensive}. To achieve optimal performance on specific tasks, these pre-trained models often require further fine-tuning. However, conventional fine-tuning methods are typically centralized, involving the collection of raw data on a single client, which raises significant privacy concerns. 
Federated Learning (FL), as a privacy-preserving distributed Machine Learning (ML) training paradigm\cite{mcmahan2017communication,zhang2021survey,wen2023survey}, is a promising approach to address this issue by enabling collaborative fine-tuning across decentralized LLM agents without aggregating raw datasets. Integrating FL with LLMs not only addresses privacy concerns but also leverages diverse data sources from decentralized clients to fine-tune LLMs on specific domains.

As FL increasingly relies on wireless networks to support collaboration across clients, communication has emerged as a critical bottleneck in fine-tuning LLMs. Unlike centralized settings, FL requires frequent transmission of model updates over limited-bandwidth connections, posing significant challenges for models with billions of parameters, such as LLaMA3-8B\cite{dubey2024llama}. Even in half-precision, transmitting these parameters demands approximately 16 GB per update, resulting in delays that significantly impact training efficiency. Moreover, the large parameter size imposes high demands on the client's computational capabilities. For instance, full fine-tuning of LLaMA3-8B requires memory for model parameters, optimizer states (twice the size of the parameters), and gradients (equivalent to the size of the parameters). Even with mixed precision, it requires at least 64 GB of GPU memory per client, posing significant challenges for resource-constrained devices.

To address these challenges, LLM quantization \cite{zhou2024survey, zhu2024survey,kim2024efficient} has been proposed as a solution to alleviate computational resource constraints, such as GPU memory usage, in federated LLM fine-tuning. The authors in \cite{fang2024automated} proposed FedPipe, which applied low-precision quantization to reduce computational resource consumption. However, this approach employs a uniform quantization strategy across all clients, overlooking the potential heterogeneity in their computational capabilities. In practice, clients with powerful GPUs can support full-precision LLM fine-tuning without quantization, while aggressive quantization on such devices may lead to unnecessary performance loss. Conversely, resource-constrained clients with limited GPU memory or computational power can selectively apply higher levels of quantization to adapt to their resource limitations, highlighting the need for more adaptive strategies.

In addition to quantization, Parameter-Efficient Fine-Tuning (PEFT) methods have been proposed to further alleviate the computational and communication challenges in federated LLM fine-tuning \cite{han2024parameter,xu2023parameter,zhang2025parameter}. 
One of the widely used PEFT methods is Low-Rank Adaptation (LoRA), which freezes the original pre-trained parameters of the LLM and trains a smaller number of additional parameters instead\cite{hu2021lora}. This approach significantly reduces computational and storage requirements while maintaining a model performance comparable to full fine-tuning.
The authors in \cite{zhang2024towards} proposed an approach that integrates FL with LoRA to enhance the instruction tuning of LLMs. This method enables clients to only train and optimize the LoRA layers using locally diverse instruction data, reducing privacy exposure 
risks by sharing only the LoRA weights, which significantly lowers the demands on communication and client computational capabilities. However, the study assumed a uniform LoRA rank across all clients and does not consider the potential impact of client resource heterogeneity.

To accommodate heterogeneous computational capabilities, the authors in \cite{cho2023heterogeneous} proposed HETLoRA that assigned different LoRA ranks to different clients. 
To facilitate the aggregation and distribution of LoRA matrices with different ranks, this work introduced a zero-padding aggregation method and a global model truncation distribution method. However, the zero-padding aggregation method can dilute the information learned by high-rank clients, and the truncation approach may lead to performance loss in the models distributed to clients.

To address the challenges posed by different LoRA ranks, the authors in \cite{jiamu2024federated} synthesized a full-size LoRA through the direct multiplication of the LoRA A and B matrices, ensuring consistent dimensions of the full-size LoRA across clients.
This full-size LoRA is then aggregated to form an updated global model, which is then decomposed using Singular Value Decomposition (SVD). The SVD components are further processed using a low-rank approximation to fit different client ranks before being distributed to the clients. However, the reconstruction method in \cite{jiamu2024federated} may lose information regarding the cross-relation across clients during aggregation, and the low-rank approximation distribution method still inevitably leads to performance loss.

On the other hand, to further reduce communication costs in federated LoRA fine-tuned LLMs, the authors in \cite{zhu2024promoting} proposed integrating a quantization strategy for LoRA parameters to adapt to bandwidth-limited scenarios. However, this approach utilized fixed quantization precision across all clients, and failed to account for clients' communication heterogeneity. For instance, clients with good channel conditions could support higher precision to accelerate convergence, whereas clients with poor channel conditions might require lower precision quantization to alleviate communication bottlenecks. Moreover, their methods did not consider the importance of individual rows and columns in the LoRA A and B matrices, treating all parameters equally during transmission. This lack of prioritization can lead to inefficient communication, as important components may be lost due to bandwidth constraints, potentially slowing down convergence and degrading global model performance.

To address the limitations of previous work \cite{zhang2024towards,cho2023heterogeneous,jiamu2024federated,zhu2024promoting,fang2024automated}, we propose a novel Heterogeneous Adaptive Federated LoRA Fine-tuned LLM with Quantization (HAFLQ) framework to handle the challenges posed by heterogeneous client computational and communication resources.
Our contributions can be summarized as follows:
\begin{itemize}
\item \textbf{Salience-Driven Adaptive Quantization for Efficient LLM Deployment.} 
To enable efficient deployment of LLMs on heterogeneous devices, we propose a salience-driven and resource-aware adaptive quantization scheme. By ranking transformer blocks based on their contribution to model performance, our method applies higher precision to more important blocks while quantizing less important ones. Additionally, the quantization strategy is dynamically adapted to the computational capabilities and memory constraints of each client device. Clients with more computational resources retain more blocks in full precision, while resource-constrained clients apply lower-precision quantization to a larger proportion of blocks.

\item \textbf{Importance-Aware LoRA Fine-Tuning with Truncation and Freezing.} 
To accommodate client resource heterogeneity, we propose an importance-based parameter truncation scheme, which allows clients to fine-tune LoRA layers with different ranks based on their computational capabilities. To avoid performance degradation caused by truncation, we further develop an importance-based parameter freezing scheme. In this approach, both the cloud server and the clients maintain the same high LoRA rank, while clients selectively update only the most important LoRA rank-1 matrices while keeping others frozen.

\item \textbf{Importance-Aware Bandwidth-Adaptive Communication Quantization.} 
To address communication heterogeneity in wireless FL systems, we propose an Importance-Aware Bandwidth-Adaptive Communication Quantization scheme. This method dynamically adjusts the precision of the transmitted parameters based on their importance and each client’s bandwidth budget. By prioritizing the transmission of the most important LoRA rank-1 matrices with higher precision, the scheme optimizes bandwidth utilization and improves FL performance under constrained communication conditions.

\item \textbf{Adaptive Rank-1 Matrix-Level Model Aggregation.} 
To improve global model aggregation, we propose an adaptive aggregation method at the rank-1 matrix level. Instead of aggregating entire LoRA matrices, only the clients that update and upload specific rank-1 matrices participate in the aggregation process. This prevents information dilution and accelerates convergence compared to traditional zero-padding methods.

\item We evaluate our methods on the Banking 77 text classification task. The salience-driven adaptive LLM quantization method reduces memory usage by 31\% and computational costs, enabling efficient deployment on heterogeneous clients without sacrificing performance. Additionally, our importance-based partial freezing scheme preserves client model accuracy better than truncation-based methods, while the adaptive aggregation method achieves faster convergence compared to the zero-padding approach in \cite{cho2023heterogeneous}. Finally, our Importance-Aware Bandwidth-Adaptive Quantization optimizes bandwidth usage, reducing communication cost by 49\%, while maintaining high accuracy under bandwidth constraints.

\end{itemize}
\begin{figure*}[]
\centerline{\includegraphics[scale=0.46]{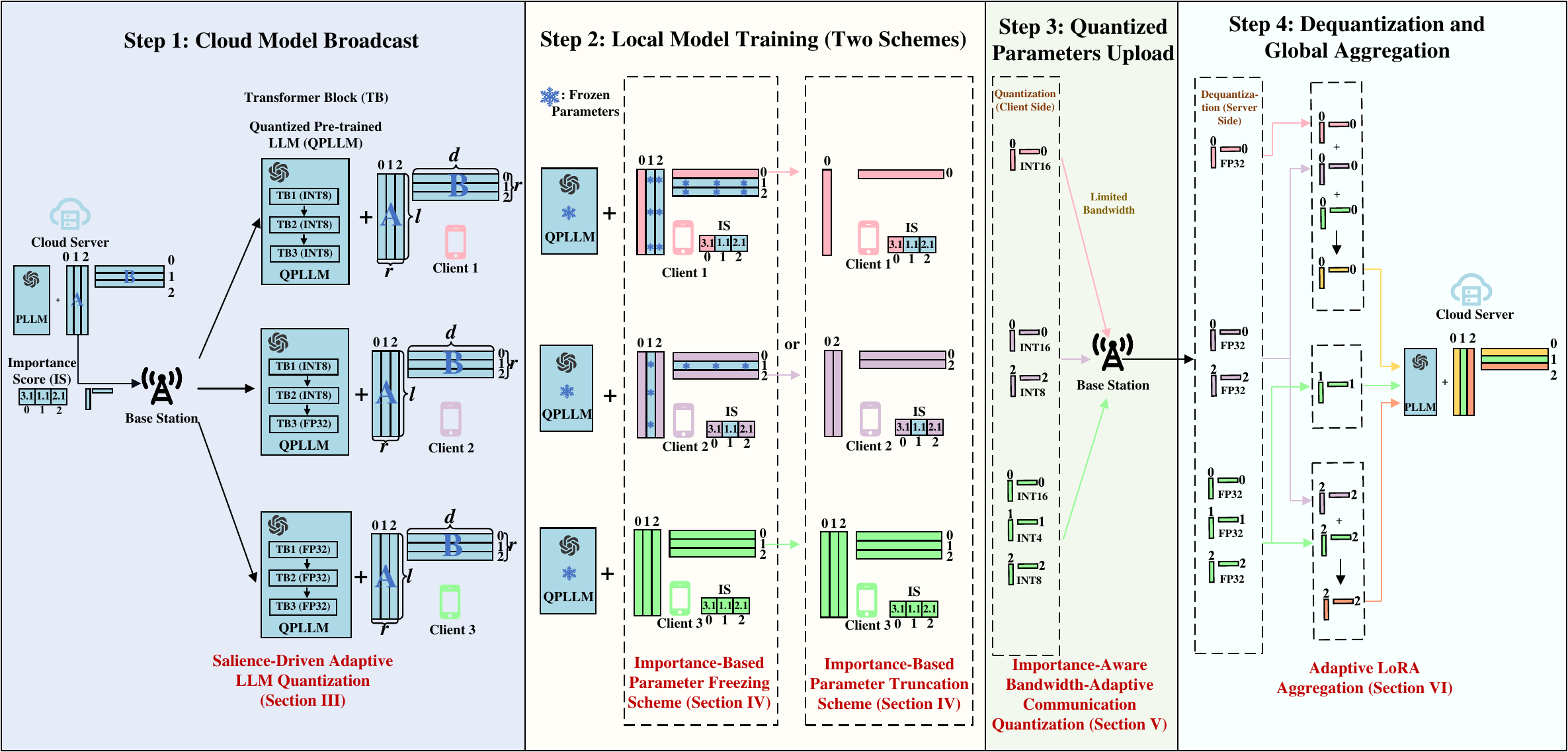}}
\caption{HAFLQ framework.}
\label{System_Model}
\end{figure*}

The rest of the paper is organized as follows:
Section II introduces the system model. Sections III and IV describe the salience-driven adaptive LLM quantization scheme and the importance-based parameter truncation and freezing schemes, respectively. Section V proposes an importance-aware and bandwidth-adaptive communication quantization scheme, followed by Section VI, which presents the adaptive global aggregation scheme. Numerical results are discussed in Section VII, and conclusions are summarized in Section VIII.

\section{System Model}
As shown in Fig.~\ref{System_Model}, our proposed HAFLQ framework consists of a cloud server, a base station, and a set of $K$ clients, denoted as $\mathcal{K} = \{1, 2, \cdots, K\}$. Each client and the cloud server have a LoRA based LLM, where the parameters of the original pre-trained LLM are frozen, and only the incorporated low-rank decomposition matrices are trainable. Each client $k \in \mathcal{K}$ has a local dataset $\mathcal{D}_k = \{(x_i, y_i)\}_{i=1}^{D_k}$, where $x_i$ is the $i$th input data sample, $y_i$ is the corresponding labeled output, and $D_k$ is the number of data samples. 
Each client connects to the base station via wireless channels to transmit parameters to the cloud server.

\subsection{Cloud LoRA Parameter Broadcast}
The updated model parameters of the LoRA-based LLM are handled differently at the cloud server and the clients to address their respective computational and memory constraints. Specifically, the cloud server retains the pre-trained model weights \(\mathbf{W}_{\text{pre}}\in \mathbb{R}^{d \times l}\) in full precision, while the clients use a quantized version of the pre-trained weights, denoted as \(\mathbf{W}_{\text{pre,q}}\in \mathbb{R}^{d \times l}\). For the cloud server, the model parameters can be expressed as
\begin{equation}
\mathbf{W}_{\text{server}} = \mathbf{W}_{\text{pre}} + \Delta \mathbf{W} = \mathbf{W}_{\text{pre}} + \mathbf{B}\mathbf{A}.
\end{equation}

For the client, the model parameters can be expressed as
\begin{equation}
\mathbf{W}_{\text{client}} = \mathbf{W}_{\text{pre,q}} + \Delta \mathbf{W} = \mathbf{W}_{\text{pre,q}} + \mathbf{B}\mathbf{A},
\end{equation}
where \(\Delta \mathbf{W} \in \mathbb{R}^{d \times l}\) is the trainable matrix. The low-rank decomposition of \(\Delta \mathbf{W}\) is given by \(\Delta \mathbf{W} = \mathbf{B}\mathbf{A}\), where \(\mathbf{B} \in \mathbb{R}^{d \times r}\) and \(\mathbf{A} \in \mathbb{R}^{r \times l}\) are low-rank matrices. Here, \(d\) and \(l\) are the dimensions of the model's weight matrix, and \(r\) is the rank of the low-rank approximation, which is typically much smaller than both \(d\) and \(l\).

For different clients \(k\), the LoRA rank is denoted as \(r_k\). In our importance-based parameter truncation scheme, the client LoRA rank \(r_k\) is selected from the range \([r_{\text{min}}, r_{\text{max}}]\). In our proposed importance-based parameter freezing scheme, rank \(r_k\) is set to be equal to \(r_{\text{max}}\). Regardless of the scheme used, the LoRA rank of the global model at the cloud server is denoted as \(r_g\), where \(r_g = r_{\text{max}}\).

During each communication round, the cloud server broadcasts the LoRA parameters, which are the low-rank matrices \(\mathbf{B}_g\) and \(\mathbf{A}_g\), to all clients \(k \in \mathcal{K}\). Taking the communication round \(t\) as an example, these parameters are denoted as
\begin{equation}
\Theta_{\text{LoRA}}^{(t)} = \{\mathbf{B}_g^{(t)},\mathbf{A}_g^{(t)}\},
\end{equation}
where \(\mathbf{B}_g^{(t)} \in \mathbb{R}^{d \times r_g}\) and \(\mathbf{A}_g^{(t)} \in \mathbb{R}^{r_g \times l}\).

\subsection{Local Model Training and Global Aggregation}
In each communication round, the selected set of clients \(\mathcal{K}\) receives the LoRA parameters \(\Theta_{\text{LoRA}}^{(t)}\) broadcasted by the cloud server. Each client \(k \in \mathcal{K}\) then begins local model training using its local dataset \(\mathcal{D}_k\). In this process, clients keep the pre-trained model weights \(\mathbf{W}_{\text{pre}}\) frozen. They initialize their local LoRA parameters \(\mathbf{B}_k^t\) and \(\mathbf{A}_k^t\) with the received \(\Theta_{\text{LoRA}}^{(t)}\) and optimize these parameters by minimizing a local loss function.

After completing local training, clients adaptively quantize their updated LoRA parameters, \(\{\mathbf{B}_k^{(t+1)}, \mathbf{A}_k^{(t+1)}\}\), into \(\{\mathbf{B}_{k,\text{q}}^{(t+1)}, \mathbf{A}_{k,\text{q}}^{(t+1)}\}\) before uploading them to the cloud server. Unlike traditional fixed or uniform quantization methods, our proposed importance-aware bandwidth-adaptive quantization adjusts the precision of each parameter based on its importance and the client’s bandwidth constraints. This ensures that the most critical rank-1 matrices are prioritized for transmission with higher precision, optimizing bandwidth utilization and minimizing performance degradation in heterogeneous environments.

Upon receiving updates, the cloud server dequantizes the parameters back to \(\{\mathbf{B}_{k,\text{dq}}^{(t+1)}, \mathbf{A}_{k,\text{dq}}^{(t+1)}\}\) and aggregates them using our proposed Adaptive Rank-1 Matrix-Level Aggregation method. This approach aggregates only the updated and important rank-1 matrices, avoiding information dilution and accelerating convergence compared to traditional zero-padding methods. These innovations enable the HAFLQ framework to achieve efficient communication, scalability, and high performance in federated learning.

\subsection{Wireless Communication Model}
\subsubsection{Uplink Communication}
In our HAFLQ framework, each communication round is of fixed duration $T$ and each client $k$ is allocated a limited bandwidth $B_k$. Consequently, the number of bits that a client can transmit in a given communication round is constrained.
To characterize the wireless uplink between a client and the base station, we adopt a channel model that accounts for path loss, shadowing, and small-scale fading.

\paragraph{OFDMA-based Multi-User Access}  
When multiple clients are scheduled to transmit concurrently, an orthogonal frequency division multiple access (OFDMA) scheme is adopted to avoid inter-client interference. We let $\mathcal{C}$ denote the set of available subchannels, where each subchannel has identical bandwidth $B_0$. Then, a binary channel assignment vector $\boldsymbol{\rho}_k = [\rho_{k,1}, \rho_{k,2}, \ldots, \rho_{k,|\mathcal{C}|}]$ can be defined for client $k$, where $\rho_{k,c}=1$ indicates that subchannel $c$ is allocated to client $k$. The constraints 
\begin{align}
    \sum_{c\in\mathcal{C}} \rho_{k,c} &= 1, \quad \forall k \in \mathcal{K}, \label{eq:assign1}\\
    \sum_{k\in\mathcal{K}} \rho_{k,c} &\le 1, \quad \forall c \in \mathcal{C}, \label{eq:assign2}
\end{align}
ensure that each client is assigned exactly one subchannel and each subchannel is shared by at most one client in a given round.
Since all subchannels have equal bandwidth \( B_0 \), the bandwidth allocated to client \( k \) is \( B_k = B_0 \) for all \( k \in \mathcal{K} \).

\paragraph{Channel Gain and Path Loss}  
We consider a common urban scenario where client \( k \) has no line-of-sight (NLoS) to the base station\cite{3gpp_tr_38_901_v16_1_0} and the path loss is modeled as  
\begin{equation}
    PL_k = 32.4 + 20\log_{10}(f) + 30\log_{10}(d_k),
    \label{eq:pl}
\end{equation}
where \( f \)  is the carrier frequency and \( d_k \) is the distance between client \( k \) and the base station.  

Incorporating shadowing and small-scale fading effects, the overall channel gain \( h_k \) is given by  
\begin{equation}
    h_k = 10^{-PL_k/10} \, \psi_k \, \chi_k,
    \label{eq:channel_gain}
\end{equation}
where the large-scale shadowing effect is modeled as  
\begin{equation}
\psi_k = 10^{\xi_k/10}, \quad \xi_k \sim \mathcal{N}(0, \sigma^2),
\end{equation}
and the small-scale Rayleigh fading component is  
\begin{equation}
\chi_k = |g|^2, \quad \text{with } g \sim \mathcal{CN}(0,1).
\end{equation}

Here, \(\mathcal{N}(0, \sigma^2)\) denotes a real-valued Gaussian distribution with zero mean and variance \(\sigma^2\), and \(\mathcal{CN}(0,1)\) denotes a circularly symmetric complex Gaussian distribution with zero mean and unit variance.

\paragraph{Signal-to-Noise Ratio and Data Rate}  
Assuming that each client transmits with power $P_k$, the Signal-to-Noise Ratio (SNR) at the base station for client $k$ is given by
\begin{equation}
    \gamma_k = \frac{P_k\, h_k}{N_0^{\text{bs}}\, B_k},
    \label{eq:snr}
\end{equation}
where $N_0^{\text{bs}}$ is the noise power spectral density (PSD) at the base station side. Under the Shannon capacity formulation, the achievable uplink data rate\cite{7892899} of client $k$ is
\begin{equation}
    R_k = B_k \log_2\left(1 + \gamma_k\right).
    \label{eq:rate}
\end{equation}

Thus, during the uplink phase of a communication round, which lasts for $T_{\text{ul}}$, the total number of bits that client $k$ can transmit is
\begin{equation}
    M_k = T_{\text{ul}} R_k.
    \label{eq:bits}
\end{equation}

\paragraph{Parameter Transmission}  
In each round, client $k$ transmits its updated LoRA parameters, which together occupy $S_k$ bits. Since the available bit budget per round is limited by~\eqref{eq:bits}, we require
\begin{equation}
    S_k \le M_k,
    \label{eq:bit_constraint}
\end{equation}
for successful transmission. 

\subsubsection{Downlink Communication}
In the proposed HAFLQ framework, we assume that the base station broadcasts the global LoRA parameters \(\Theta_{\text{LoRA}}^{(t)} = \{\mathbf{B}_g^{(t)}, \mathbf{A}_g^{(t)}\}\) to all clients in full precision using shared channels. Given the low-rank nature of LoRA parameters, their dimensions are significantly smaller compared to the full model parameters, resulting in low communication overhead. Additionally, the high transmit power of the base station ensures that these parameters can be reliably delivered to all clients, even under varying channel conditions.

\section{Salience-Driven Adaptive LLM Quantization}
In wireless FL systems, clients often have varying computational capabilities due to differences in hardware resources such as GPU memory and processing power. Deploying LLMs across such heterogeneous clients requires computation resource optimization, particularly GPU memory and computation.

To address this, we propose an adaptive quantization approach for LLMs that customizes the quantization level based on each client's computational capability. Our method leverages a salience metric to assess the importance of transformer blocks, enabling selective quantization. Important transformer blocks, identified as having higher salience, are preserved in full precision to maintain model performance, while less important blocks are quantized to lower precision.  
The proportion of blocks quantized depends on the GPU memory available to the client. Clients with limited GPU memory quantize a larger proportion of transformer blocks to fit within their memory constraints, while clients with greater GPU memory retain more blocks in full precision to maximize model performance.

\subsection{Salience-Driven LLM Transformer Block Importance Evaluation}
In our approach, the pre-trained LLM is composed of \( L \) transformer blocks that are stacked sequentially. Each transformer block \( l \) has parameters \( \mathbf{W}_{\text{TB}_l} \in \mathbb{R}^{d \times d} \). To evaluate the importance of each transformer block, we adopt a salience-based metric inspired by SparseGPT~\cite{DBLP:conf/icml/FrantarA23}, which quantifies the contribution of each weight element to the model's output error.
Specifically, for the weight matrix \( \mathbf{W}_{\text{TB}_l} \) in a given transformer block, we compute the salience value \( \delta_{i,j} \) for each element \( w_{i,j} \) using
\begin{equation}
\delta_{i,j} = \frac{w_{i,j}^2}{[\mathbf{H}^{-1}]_{j,j}}
\end{equation}
where \( [\mathbf{H}^{-1}]_{j,j} \) is the \( j \)th diagonal element of inverse Hessian matrix \( \mathbf{H}^{-1} \). 
The Hessian matrix \( \mathbf{H} \in \mathbb{R}^{d \times d} \) is computed as the average outer product of input activation vectors \( \mathbf{x}_i \) in the local dataset \( \mathcal{D}_k \)
\begin{equation}
\mathbf{H} = \frac{1}{D_k} \sum_{i=1}^{D_k} \mathbf{x}_i \mathbf{x}_i^T,
\end{equation}
where \( \mathbf{x}_i \) is the input activation vector for the \( i \)th sample in the dataset \( \mathcal{D}_k \). To efficiently compute the diagonal elements of \( \mathbf{H}^{-1} \), we leverage Cholesky decomposition~\cite{6710599}, which avoids explicitly inverting \( \mathbf{H} \) and allows for efficient computation of the required diagonal elements.

Next, we calculate the total salience score for each transformer block \( \mathbf{W}_{\text{TB}_l} \) by summing the salience values \( \delta_{i,j} \) of all its weight elements
\begin{equation}
R_{\text{TB}_l} = \sum_{i=1}^d \sum_{j=1}^d \delta_{i,j},
\end{equation}
the salience score \( R_{\text{TB}_l} \) represents the overall impact of the transformer block $l$ on the model's output error. Based on these salience scores, we rank all transformer blocks in descending order of importance.

This salience-driven evaluation provides the foundation for our quantization strategy. During quantization, transformer blocks with lower salience scores are prioritized for quantization, while those with higher salience scores are preserved in full precision to minimize the impact on model performance. 
The proportion of blocks to be quantized is dynamically adjusted based on the computational capacity and GPU memory size of the target device: for devices with higher computational power and larger GPU memory, only a small number of blocks are quantized, preserving most blocks in full precision. Conversely, for devices with lower computational power and smaller GPU memory, a larger proportion of blocks are quantized to meet resource constraints.

\subsection{Block-wise Quantization for LLMs}
To efficiently quantify LLMs while minimizing the loss of accuracy, we utilize block-wise quantization \cite{tim2022llm,frantar2022gptq}. By dividing the large weight matrices into smaller blocks and quantizing each block independently, this method allows the quantization process to better adapt to the local distribution of weights.

\subsubsection{Divide the Weight Matrix}
Given a weight matrix \( \mathbf{W}_{\text{TB}_l} \in \mathbb{R}^{d \times d} \) from transformer block \( l \), which is selected for quantization, the matrix is divided into several contiguous blocks along the row dimension. Specifically, assuming a block size of \( m \), the matrix is partitioned into \( p = d / m \) blocks: \( \mathbf{W}_1, \mathbf{W}_2, \dots, \mathbf{W}_p \), where each block \( \mathbf{W}_i \in \mathbb{R}^{m \times d} \) contains \( m \) consecutive rows of the original matrix.

\subsubsection{Quantize the Block}
For each block \( \mathbf{W}_i \), a scale factor \( s_i \) is computed based on the range of values within the block using
\begin{equation}
s_i = \frac{\max(\mathbf{W}_i) - \min(\mathbf{W}_i)}{2^q - 1},
\end{equation}
where \( q \) denotes the number of quantization bits. The quantized weights for block \( \mathbf{W}_i \) are then obtained as
\begin{equation}
\mathbf{W}_{i,\text{q}} = \left\lfloor \frac{\mathbf{W}_i}{s_i} \right\rceil,
\end{equation}
where \( \left\lfloor \cdot \right\rceil \) denotes the rounding operation to the nearest integer.

The quantized weights are stored in a compressed format in the device's GPU memory, to ensure efficient access during computation. This storage format significantly reduces memory usage compared to full-precision weights, enabling the deployment of large models on resource-constrained devices.

\subsubsection{Dequantize the Block} 
The dequantization process occurs dynamically during the forward pass, both in training and inference. Specifically, when a quantized weight block \( \mathbf{W}_{i,\text{q}} \) is needed for computation, it is dequantized on-the-fly using the corresponding scale \( s_i \) as
\begin{equation}
\mathbf{W}_i = s_i \cdot \mathbf{W}_{i,\text{q}}.
\end{equation}

This approach ensures that the computational overhead of dequantization is minimized in scenarios where only the required blocks are dequantized on-the-fly during computation, rather than the entire weight matrix.

\section{Importance-based Parameter Truncation and Freezing Schemes}
In this section, to handle the heterogeneity of client resources, we first introduce an importance-based parameter truncation scheme at the local LLMs. To address the issue of performance degradation resulting from truncation, we subsequently introduce an importance-based parameter freezing scheme at the local LLMs.

\subsection{Decomposed LoRA Rank-1 Matrix Importance Evaluation}
In practical FL environments, clients often have different computational capacities, making it challenging to train all clients' LoRA parameters with a uniform rank across all clients. To address this, the LoRA parameter product \(\mathbf{B}\mathbf{A}\) is decomposed into smaller, rank-1 matrices. As shown in Fig.~\ref{System_Model}, we take LoRA with $r_g=3$ as an example. The cloud server calculates an importance score for each rank-1 matrix and broadcasts this information to the clients. Clients can then prioritize these rank-1 matrices, training the most important ones first.

\subsubsection{Matrix Decomposition}
The cloud server decomposes the product \(\mathbf{B}_g\mathbf{A}_g\) into rank-1 matrices as
\begin{equation}
\mathbf{B}_g\mathbf{A}_g = \sum_{i=1}^{r_g} \mathbf{b}_i \mathbf{a}_i,
\end{equation}
where \(\mathbf{b}_i \in \mathbb{R}^{d\times1}\) is the \(i\)th column vector of \(\mathbf{B}_g\), and \(\mathbf{a}_i \in \mathbb{R}^{1 \times l}\) is the \(i\)th row vector of \(\mathbf{A}_g\).

\subsubsection{Element-wise Importance Calculation}
To assess the significance of each element within vectors \(\mathbf{b}_i\) and \(\mathbf{a}_i\), we employ a sensitivity-based metric\cite{Molchanov_2019}. This involves computing the gradient of the global loss function \(L\) with respect to each element and using the product of the element's value and its gradient as a measure of importance. However, in the context of FL, the global model does not have a direct loss function $L$ to calculate the gradient, as the training data is distributed across multiple clients, and the global model is updated based on aggregated client updates. Instead, we approximate the gradient by scaling the change in parameter values between consecutive communication rounds using the learning rate.
Then, we define a unified importance function \(I(w_{ij})\) for any trainable parameter \(w_{ij}\) as
\begin{equation}
I(w_{ij}) = \left| w_{ij} \cdot \frac{\Delta w_{ij}}{\eta} \right|,
\end{equation}
where \(w_{ij}\) represents element \(b_{ji}\) in \(\mathbf{b}_i\), with \(j \in \{1, \ldots, d\}\), or element \(a_{iq}\) in \(\mathbf{a}_i\), with \(q \in \{1, \ldots, l\}\). The term \(\Delta w_{ij} = w_{ij}^{(t)} - w_{ij}^{(t-1)}\) represents the change in the parameter value between two consecutive global updates, and $\eta$ is the learning rate.

However, due to the high variability and uncertainty in estimating sensitivity from mini-batch samples, it is crucial to smooth these importance scores to obtain more reliable indicators. We achieve this by applying an exponential moving average to the sensitivity scores and quantifying the uncertainty\cite{DBLP:conf/icml/ZhangZLBHCZ22}.

First, the smoothed sensitivity \(\bar{I}^{(t)}(w_{ij})\) is calculated using an exponential moving average
\begin{equation}
\bar{I}^{(t)}(w_{ij}) = \beta_1 \bar{I}^{(t-1)}(w_{ij}) + (1 - \beta_1) I^{(t)}(w_{ij}),
\end{equation}
where \(\beta_1 \in [0,1]\) is a smoothing factor.

Next, the uncertainty \(\bar{U}^{(t)}(w_{ij})\) is quantified to capture the variability in the sensitivity scores
\begin{equation}
\bar{U}^{(t)}(w_{ij}) = \beta_2 \bar{U}^{(t-1)}(w_{ij}) + (1 - \beta_2) \left| I^{(t)}(w_{ij}) - \bar{I}^{(t)}(w_{ij}) \right|,
\end{equation}
where \(\beta_2 \in [0,1]\) is another smoothing factor.

Finally, the combined importance score \(s^{(t)}(w_{ij})\) is defined as the product of the smoothed sensitivity and the uncertainty:
\begin{equation}
s^{(t)}(w_{ij}) = \bar{I}^{(t)}(w_{ij}) \cdot \bar{U}^{(t)}(w_{ij}).
\end{equation}
By incorporating these smoothing techniques, we enhance the reliability of the importance scores, making them more robust to the stochasticity of FL environments.

\subsubsection{Aggregation of Importance Scores}
To obtain the overall importance score for rank-1 matrix \(\mathbf{b}_i \mathbf{a}_i\), we sum the smoothed and uncertainty adjusted importance scores for all elements of \(\mathbf{b}_i\) and \(\mathbf{a}_i\). This approach ensures that the aggregated score reflects the reliability and significance of each parameter. The overall importance score \(S_i\) of the rank-1 matrix is calculated as
\begin{equation}
S_i = \sum_{j=1}^{d} s(b_{ji}) + \sum_{q=1}^{l} s(a_{iq}),
\end{equation}
where \(s(b_{ji})\) and \(s(a_{iq})\) are the smoothed and uncertainty-adjusted importance scores for the elements of \(\mathbf{b}_i\) and \(\mathbf{a}_i\), respectively. This aggregation method provides a robust measure of the importance of each rank-1 matrix in the decomposition of \(\mathbf{B}\mathbf{A}\).

\subsubsection{Broadcast Importance Scores}
After calculating the importance scores \(S\) for each rank-1 matrix, we obtain a list of scores \(\left[S_1, S_2, \ldots, S_{r_g}\right]\). This list represents the importance of each rank-1 matrix in the decomposition of \(\mathbf{B}\mathbf{A}\). The list of importance scores is then broadcast to all clients.

\subsection{Importance-based Parameter Truncation Scheme}
To accommodate diverse computational capabilities of clients, we allow each client to have a different LoRA rank \(r_k\in[r_{\text{min}}, r_{\text{max}}]\). However, this flexibility introduces the challenge of aligning the cloud server's global model with the different LoRA ranks of each client’s local model. To address this issue, we propose an importance-based parameter truncation scheme.

\subsubsection{Truncation Process}
In each communication round $t$, the cloud server broadcasts the LoRA parameters \(\Theta_{\text{LoRA}}^{(t)}\) to all clients. Upon receiving these parameters, each client \(k \in \mathcal{K}\) performs the following steps.
\paragraph{Selection of Significant Rank-1 Matrices}
Each client \(k \in \mathcal{K}\) receives the list of importance scores \([S_1, S_2, \ldots, S_{r_g}]\) from the server. Based on these scores, client \(k\) selects the top \(r_k\) rank-1 matrices as
\begin{equation}
\label{Ik_1}
\mathcal{I}_k^{\text{T}} = \text{topk}\left([S_1, S_2, \ldots, S_{r_g}], r_k\right),
\end{equation}
where $\mathcal{I}_k^{\text{T}}$ represents the set of indices corresponding to the rank-1 matrices selected by client $k$.

\paragraph{Truncation of Low-Rank Matrices} Client \(k\) then truncates the low-rank matrices \(\mathbf{B}_g\) and \(\mathbf{A}_g\) to align with the local model's LoRA rank requirement $r_k$. Specifically, the client retains \(r_k\) column vectors \(\mathbf{b}_i\) and \(r_k\) row vectors \(\mathbf{a}_i\) corresponding to the highest importance scores
\begin{equation}
\label{B_kA_k}
\begin{aligned}
\mathbf{B}_k^{\text{T}} = \mathbf{B}_g[:, \mathcal{I}_k^{\text{T}}],  \quad \mathbf{A}_k^{\text{T}} = \mathbf{A}_g[\mathcal{I}_k^{\text{T}}, :].
\end{aligned}
\end{equation}

\subsubsection{Local Model Update} The truncated low-rank matrices \(\mathbf{B}_k^{\text{T}}\) and \(\mathbf{A}_k^{\text{T}}\) are copied to the LoRA parameters of client $k$'s local model. Subsequently, client $k$ trains the local model using its local dataset $\mathcal{D}_k$.
In the parameters truncation scheme, the local loss function for each client \(k\) is defined as
\begin{align}
L_k^{\text{T}}(\mathbf{B}_k^{\text{T}}, \mathbf{A}_k^{\text{T}}) &= \frac{1}{|\mathcal{D}_k|} \sum_{\xi \in \mathcal{D}_k} 
\ell\left(\left(\mathbf{B}_k^{\text{T}}, \mathbf{A}_k^{\text{T}}\right), \xi \mid 
\mathbf{W}_{\text{pre}}\right) \notag\\
&+\frac{\lambda}{2} \left( \|\mathbf{B}_k^{\text{T}}\|^2 + \|\mathbf{A}_k^{\text{T}}\|^2 \right),
\end{align}
where \(|\mathcal{D}_k|\) denotes the size of the dataset. The function \(\ell\) measures the model’s performance on a data sample \(\xi\). Parameters \(\mathbf{B}_k^{\text{T}}\) and \(\mathbf{A}_k^{\text{T}}\) are the trainable LoRA rank-1 matrices that are being optimized.
The term \(\lambda\) is the weight decay coefficient, which is used to prevent overfitting by penalizing large parameter values. The regularization term \(\frac{\lambda}{2} \left( \|\mathbf{B}_k^{\text{T}}\|^2 + \|\mathbf{A}_k^{\text{T}}\|^2 \right)\) incorporates the L2 norm, \(\|\cdot\|\), to measure the magnitude of the parameters, thereby encouraging smaller parameter values.

\subsection{Importance-based Parameter Freezing Scheme}
While the truncation process described above offers the advantage of simple implementation, it inevitably results in performance degradation when distributing the model to clients. To address this issue, we propose a novel importance-based parameter freezing scheme, where both local models and the global model utilize the same maximum LoRA rank, denoted as \(r_k = r_g = r_{\text{max}}\). To accommodate client resource constraints, clients will freeze a portion of the LoRA parameters based on their importance. We denote the freezing ratio as $\alpha_k$, which represents the proportion of LoRA parameters that are frozen relative to all parameters.

\subsubsection{Freezing Process}
In the HAFLQ framework, each client updates its local LoRA parameters by selectively training a subset of the parameters based on the importance scores received from the cloud server. All LoRA parameters \(\Theta_{\text{LoRA}} = \{\mathbf{B}_g, \mathbf{A}_g\}\) broadcasted by the server are replicated to the client's local model, where the local LoRA parameters are initialized as \(\mathbf{B}_k^{\text{F}} = \mathbf{B}_g\) and \(\mathbf{A}_k^{\text{F}} = \mathbf{A}_g\), but only the most important rank-1 matrices are unfrozen and optimized by the client's local dataset. 
Based on the freezing ratio $\alpha_k$ of each client $k$, $(1-\alpha_k)r_{\rm max}$ rank-1 matrices will be fine-tuned.

\paragraph{Selection of Significant Rank-1 Matrices}
Each client \(k \in \mathcal{K}\) receives the list of importance scores \([S_1, S_2, \ldots, S_{r_g}]\) from the server. Based on these scores, client \(k\) selects the top \((1-\alpha_k)r_{\rm max}\) rank-1 matrices to actively train, where \(\alpha_k\) can vary between clients depending on their computational capacity and resource availability. The selection is performed by choosing the indices corresponding to the highest importance scores:
\begin{equation}
\label{Ik_2}
\mathcal{I}_k^{\text{F}} = \text{topk}\left([S_1, S_2, \ldots, S_{r_g}], (1-\alpha_k)r_{\rm max}\right),
\end{equation}
where $\mathcal{I}_k^{\text{F}}$ is the set of indices of the selected rank-1 matrices for client $k$.
The remaining rank-1 matrices' indices are denoted as \(\mathcal{U}_k^{\text{F}} = \{1, 2, \ldots, r_{\rm max}\} \setminus \mathcal{I}_k^{\text{F}}\).

\paragraph{Freezing Low-Rank Matrices}  
Client \(k\) then freezes the low-rank matrices \(\mathbf{B}_k^{\text{F}}\) and \(\mathbf{A}_k^{\text{F}}\) to align with the local model's LoRA freezing ratio requirements. Specifically, the trained LoRA parameter rank-1 matrices are retained as 
\begin{equation}
\label{B_IkA_Ik}
\begin{aligned}
\mathbf{B}_k^{\mathcal{I}} = \mathbf{B}_k^{\text{F}}[:, \mathcal{I}_k^{\text{F}}], \quad \mathbf{A}_k^{\mathcal{I}} = \mathbf{A}_k^{\text{F}}[\mathcal{I}_k^{\text{F}}, :].
\end{aligned}
\end{equation}  
Additionally, the frozen LoRA parameter rank-1 matrices are given by  
\begin{equation}
\begin{aligned}
\mathbf{B}_k^{\mathcal{U}} = \mathbf{B}_k^{\text{F}}[:, \mathcal{U}_k^{\text{F}}], \quad \mathbf{A}_k^{\mathcal{U}} = \mathbf{A}_k^{\text{F}}[\mathcal{U}_k^{\text{F}}, :].
\end{aligned}
\end{equation}  

\subsubsection{Local Model Update}
Each client \(k\) updates its local model by minimizing a loss function \(L_k^{\text{F}}\) over its dataset \(\mathcal{D}_k\). The loss function is defined as
\begin{align}
&\quad L_k^{\text{F}}(\mathbf{B}_k^{\mathcal{I}}, \mathbf{A}_k^{\mathcal{I}}) \notag \\
&= \frac{1}{|\mathcal{D}_k|} \sum_{\xi \in \mathcal{D}_k} 
\ell\left(\left(\mathbf{B}_{k}^{\mathcal{I}}, \mathbf{A}_{k}^{\mathcal{I}}\right), \xi
\;\middle|\;
\mathbf{W}_{\text{pre}}, \mathbf{B}_{k}^{\mathcal{U}}, \mathbf{A}_{k}^{\mathcal{U}}\right) \notag\\
&+\frac{\lambda}{2} \left( \|\mathbf{B}_{k}^{\mathcal{I}}\|^2 + \|\mathbf{A}_{k}^{\mathcal{I}}\|^2 \right).
\end{align}

In the importance-based parameters freezing scheme, clients receive LoRA parameters \(\{\mathbf{B}_g, \mathbf{A}_g\}\) from the cloud server and replicate all of them to their local LoRA rank-1 matrices. However, only the rank-1 matrices indexed by \(\mathcal{I}_k^{\text{F}}\) are trained and optimized, while those indexed by \(\mathcal{U}_k^{\text{F}}\) remain frozen during local training.

\section{Importance-aware Bandwidth-Adaptive Communication Quantization}
In wireless FL systems, communication overhead is constrained by limited bandwidth and channel fading, which directly impact the maximum number of bits that can be transmitted. If all parameters are transmitted in full precision, the bandwidth demand would exceed practical limits. On the other hand, static quantization with uniform precision ignores parameter importance, potentially harming model performance.

To address this challenge, we propose an Importance-Aware Bandwidth-Adaptive Communication Quantization scheme for efficient uploading of LoRA parameters. In this scheme, more important LoRA parameters are prioritized for transmission and are assigned higher precision (i.e., more bits), while less important parameters are transmitted with lower precision. The precision of each parameter is dynamically adjusted to align with the bandwidth budget $M_k$ of each client $k$, defined in (\ref{eq:bits}) in Section II, ensuring efficient utilization of the available communication resources. If the bandwidth budget is insufficient to transmit all parameters, the least important parameters may be omitted entirely from transmission. 

\subsection{Importance-Based Priority Transmission}
During each communication round, clients upload their locally trained LoRA parameters to the central cloud server. The specific parameters uploaded depend on the scheme employed. In the parameter truncation scheme, each client \(k\) uploads the LoRA parameters \(\mathbf{B}_k^{\text{T}}\) and \(\mathbf{A}_k^{\text{T}}\), as defined in (\ref{B_kA_k}). In the parameter freezing scheme, each client \(k\) uploads subsets of these parameters, denoted as \(\mathbf{B}_k^{\mathcal{I}}\) and \(\mathbf{A}_k^{\mathcal{I}}\), as defined in (\ref{B_IkA_Ik}).

Each LoRA parameter matrix can be decomposed into rank-1 matrices, represented as \(\mathbf{B}_k^{\text{T}}[:,i]\mathbf{A}_k^{\text{T}}[i,:]\) for the truncation scheme, where \(i \in [0, r_k)\), or \(\mathbf{B}_k^{\mathcal{I}}[:,i]\mathbf{A}_k^{\mathcal{I}}[i,:]\) for the freezing scheme, where \(i \in [0, (1-\alpha_k)r_{\text{max}})\). Due to the selection process \(\mathcal{I}_k^{\text{T}}\) in (\ref{Ik_1}) and \(\mathcal{I}_k^{\text{F}}\) in (\ref{Ik_2}), the rank-1 matrices for these uploaded parameters are already ordered by decreasing importance. The first rank-1 matrix (corresponding to \(i=0\)) is the most important one, with subsequent matrices decreasing in importance.

To maximize system performance under limited bandwidth, clients transmit the rank-1 matrices in the order of their importance. Specifically, transmission begins with the most important rank-1 matrix, which corresponds to the first column of \(\mathbf{B}_k^{\text{T}}\) and the first row of \(\mathbf{A}_k^{\text{T}}\) in the truncation scheme, or the first column of \(\mathbf{B}_k^{\mathcal{I}}\) and the first row of \(\mathbf{A}_k^{\mathcal{I}}\) in the freezing scheme. Subsequent rank-1 matrices are transmitted in sequence, with their importance progressively decreasing. By prioritizing the transmission of the most important parameters, the system ensures that the limited bandwidth is utilized to maximize the overall performance of the FL system.

\subsection{Importance-Aware Bandwidth-Adaptive Communication Quantization}
Building on the importance-based priority transmission strategy, we propose an importance-aware bandwidth-adaptive communication quantization approach, as detailed in \textbf{Algorithm \ref{alg:adaptive_quantization}}. This method prioritizes the transmission of important parameters with higher precision while quantizing less important parameters have lower precision, ensuring that the total transmitted bits of client $k$ do not exceed the bandwidth budget $M_k$.

\subsubsection{Quantization Levels and Bit Allocation}
The available precision levels are predefined as  
$Q = [32, 16, 8, 4],$  
representing the number of bits per parameter in the rank-1 matrix. Specifically, 32-bit represents full precision without quantization, while 16-bit, 8-bit, and 4-bit correspond to integer quantization (INT16, INT8, INT4).
The total number of available quantization levels is denoted as \( |Q| \).
For each quantization level pair \((Q[i], Q[i+1])\), the algorithm computes the bit cost per rank-1 matrix at both higher precision level  \(b_{\text{high}}\) and the lower precision level \(b_{\text{low}}\).

\subsubsection{Base Cost Calculation}
The base cost assumes all rank-1 matrices are transmitted at the lower precision \(b_{\text{low}} \cdot N_k\),
where \( N_k \) represents the total number of rank-1 matrices for client \( k \). For example, \( N_k = r_k \) in the truncation scheme or \( N_k =(1-\alpha_k)r_{\text{max}} \) in the freezing scheme.
If the bandwidth budget \(M_k\) is insufficient to cover this base cost, the algorithm skips to the next quantization level pair.

\subsubsection{High-Precision Allocation}
For the available bandwidth \(M_k\), the algorithm determines the number of rank-1 matrices \(n_h\) that can be transmitted at the higher precision \(Q[i]\), prioritizing matrices with smaller \(i\) values (higher importance). The remaining matrices \(N_k - n_h\) are transmitted at the lower precision \(Q[i+1]\). Here, the floor function \(\lfloor . \rfloor\) is used to ensure that \(n_h\) is an integer.

\subsubsection{Fallback to Lowest Precision}
If no suitable quantization level pair satisfies the bandwidth constraint, the algorithm transmits the most important rank-1 matrices at the lowest available precision \(Q[|Q|-1]\). The number of matrices transmitted is determined by the available bandwidth, and any remaining matrices are discarded if the bandwidth is insufficient.

\begin{algorithm}
\caption{Importance-Aware Bandwidth-Adaptive Communication Quantization}
\label{alg:adaptive_quantization}
\begin{algorithmic}[1]
\REQUIRE LoRA parameters $\mathbf{B}_k^{\text{T}}, \mathbf{A}_k^{\text{T}}$ for each client $k \in \mathcal{K}$ in truncation scheme or $\mathbf{B}_k^{\mathcal{I}}, \mathbf{A}_k^{\mathcal{I}}$ in freezing scheme, total number of rank-1 matrices $N_k$, element count per rank-1 matrix $e = d + l$, maximum number of transmit bits $M_k$, quantization levels $Q = [32, 16, 8, 4]$.
\FOR{$i = 0$ to $|Q| - 2$}
    \STATE $b_{\text{high}} = Q[i] \cdot e$ \quad // Bits per rank-1 matrix at higher precision
    \STATE $b_{\text{low}} = Q[i+1] \cdot e$ \quad // Bits per rank-1 matrix at lower precision
    \STATE $b_{\text{base}} = b_{\text{low}} \cdot N_k$ \quad // Cost if all at lower precision
    \IF{$M_k >= b_{\text{base}}$}
    \STATE $\delta = b_{\text{high}} - b_{\text{low}}$ \quad // Extra cost at higher precision
    \STATE $n_h = \min\left(N_k, \left\lfloor \frac{M_k - b_{\text{base}}}{\delta} \right\rfloor\right)$
    \IF{Truncation Scheme}
        \STATE Transmit the first $n_h$ rank-1 matrices, represented as $\mathbf{B}_k^{\text{T}}[:,j]\mathbf{A}_k^{\text{T}}[j,:]$ for $j \in [0, n_h)$, at $Q[i]$-bit precision, and the remaining rank-1 matrices at $Q[i+1]$-bit precision.
    \ELSIF{Freezing Scheme}
        \STATE Transmit the first $n_h$ rank-1 matrices, represented as $\mathbf{B}_k^{\mathcal{I}}[:,j]\mathbf{A}_k^{\mathcal{I}}[j,:]$ for $j \in [0, n_h)$, at $Q[i]$-bit precision, and the remaining rank-1 matrices at $Q[i+1]$-bit precision.
    \ENDIF
    \ENDIF
\ENDFOR
\STATE $b_{\text{min}} \gets Q[|Q|-1] \cdot e$
\STATE $n_{\text{trans}} \gets \left\lfloor \frac{M_k}{b_{\text{min}}} \right\rfloor$
\STATE Transmit the first $n_{\text{trans}}$ rank-1 matrices at $Q[|Q|-1]$-bit, and discard the remaining rank-1 matrices.
\end{algorithmic}
\end{algorithm}

\subsection{Quantization at Client and Dequantization at Server}
In our proposed approach, each LoRA rank-1 matrix is represented as \(\mathbf{b}_i \mathbf{a}_i\), where \(\mathbf{b}_i \in \mathbb{R}^{d \times 1}\) and \(\mathbf{a}_i \in \mathbb{R}^{1 \times l}\). Depending on the parameter uploading scheme, these rank-1 matrices correspond to either \(\mathbf{B}_k^{\text{T}}[:,i]\mathbf{A}_k^{\text{T}}[i,:]\) in the truncation scheme or \(\mathbf{B}_k^{\mathcal{I}}[:,i]\mathbf{A}_k^{\mathcal{I}}[i,:]\) in the freezing scheme. To reduce communication costs, we quantize vectors \(\mathbf{a}_i\) and \(\mathbf{b}_i\) independently using their respective scale and zero point parameters\cite{jacob2018quantization}. The quantized vectors and their parameters are then transmitted to the server for dequantization.

\subsubsection{Quantization at Client}
For a given vector \(\mathbf{v}\) (which can be either \(\mathbf{a}_i\) or \(\mathbf{b}_i\)), the quantization process is as follows:

\paragraph{Calculate Scale and Zero Point}
For a specified bit quantization precision \(q\), the scale $s_v$ and zero point $z_v$ are computed as follows:
\begin{align}
    s_v &= \frac{\max(\mathbf{v}) - \min(\mathbf{v})}{2^q - 1}, \\
    z_v &= \left\lfloor-\frac{\min(\mathbf{v})}{s_v}\right\rceil,
\end{align}
where \(\max(\mathbf{v})\) and \(\min(\mathbf{v})\) represent the maximum and minimum values of the elements in the vector \(\mathbf{v}\), respectively.

\paragraph{Quantize the Vector}
    \begin{equation}
    \mathbf{v}_\text{q} = \text{clamp}\left(\left\lfloor\frac{\mathbf{v}}{s_v} + z_v\right\rceil,0, 2^q - 1\right),
    \end{equation}
where clamp ($\cdot$) constrains the value within the integer range. Note that the scale \(s_v\) and zero point \(z_v\) are scalar values applied element-wise to the vector \(\mathbf{v}\), which means each element of \(\mathbf{v}\) is quantized independently.

This process is applied separately to \(\mathbf{b}_i\) and \(\mathbf{a}_i\), resulting in quantized vectors \(\mathbf{b}_{i,\text{q}}\) and \(\mathbf{a}_{i,\text{q}}\), along with their respective parameters \(s_{b_i}, z_{b_i}\) and \(s_{a_i}, z_{a_i}\). The quantized vectors, together with their respective scales and zero points, are then transmitted to the cloud server.

\subsubsection{Dequantization at Cloud Server}
Upon receiving the quantized vectors and respective scales and zero points, the server performs dequantization first to recover approximations of the original vectors:
\begin{align}
\mathbf{b}_{i, \text{dq}} &= s_{b_i} \cdot (\mathbf{b}_{i, \text{q}} - z_{b_i}), \\
\mathbf{a}_{i, \text{dq}} &= s_{a_i} \cdot (\mathbf{a}_{i, \text{q}} - z_{a_i}),
\end{align}
where the dequantization is applied to each quantized rank-1 matrix \(\mathbf{b}_{i, \text{q}}\) and \(\mathbf{a}_{i, \text{q}}\) using their respective scaling factors \(s_{a_i}, s_{b_i}\) and zero points \(z_{a_i}, z_{b_i}\).

\section{Adaptive Global Aggregation at the Cloud} 
After dequantization, the server collects the uploaded LoRA parameters from all clients and aggregates them to form a new global model. 
For the truncation-based scheme, the upload dequantized LoRA parameters are $\mathbf{B}_{k,\text{dq}}^{\text{T}}$ and $\mathbf{A}_{k,\text{dq}}^{\text{T}}$. In contrast, for the importance-based parameter freezing scheme, the upload dequantized LoRA parameters are $\mathbf{B}_{k,\text{dq}}^{\mathcal{I}}$ and $\mathbf{A}_{k,\text{dq}}^{\mathcal{I}}$.

\begin{algorithm}
\caption{Adaptive Aggregation of LoRA Parameters}
\label{alg:adaptive_aggregation}
\begin{algorithmic}[1]
\REQUIRE Dequantized LoRA parameters $\mathbf{B}_{k,\text{dq}}^{\text{T}}, \mathbf{A}_{k,\text{dq}}^{\text{T}}$ for each client $k \in \mathcal{K}$ in parameter truncation scheme or Dequantized LoRA parameters $\mathbf{B}_{k,\text{dq}}^{\mathcal{I}}, \mathbf{A}_{k,\text{dq}}^{\mathcal{I}}$ in parameter freezing scheme.
\STATE Initialize $\mathbf{B}_g^{(t+1)}$ and $\mathbf{A}_g^{(t+1)}$ to zero matrices; Initialize total norm list $Z$ with zero value
\FOR{$k$ in $\mathcal{K}$}
\FOR{each $j$ in $\mathcal{I}_k$}
    \STATE $Z[j]\mathrel{+}=z_k$ \quad// Equivalent to $Z[j] = Z[j] + z_k$
\ENDFOR
\ENDFOR
\FOR{$k$ in $\mathcal{K}$}
\STATE Index $i=0$
\FOR{each $j$ in $\mathcal{I}_k$}
    \IF{Truncation Scheme}
        \STATE $\mathbf{B}_g^{(t+1)}[:,j] \mathrel{+}= z_k\mathrel{/}Z[j]\cdot\mathbf{B}_{k,\text{dq}}^{\text{T}}[:,i]$
        \STATE $\mathbf{A}_g^{(t+1)}[j,:] \mathrel{+}= z_k\mathrel{/}Z[j]\cdot\mathbf{A}_{k,\text{dq}}^{\text{T}}[i,:]$
    \ELSIF{Freezing Scheme}
        \STATE $\mathbf{B}_g^{(t+1)}[:,j]\mathrel{+}=z_k\mathrel{/}Z[j]\cdot\mathbf{B}_{k,\text{dq}}^{\mathcal{I}}[:,i]$
        \STATE $\mathbf{A}_g^{(t+1)}[j,:]\mathrel{+}=z_k\mathrel{/}Z[j]\cdot\mathbf{A}_{k,\text{dq}}^{\mathcal{I}}[i,:]$
    \ENDIF
    \STATE $i=i+1$
\ENDFOR
\ENDFOR
\STATE Broadcast $\mathbf{B}_g^{(t+1)}, \mathbf{A}_g^{(t+1)}$ to all clients

\end{algorithmic}
\end{algorithm}

\paragraph{Adaptive Aggregation of LoRA Parameters}
The adaptive aggregation is shown in \textbf{Algorithm \ref{alg:adaptive_aggregation}}. Rather than aggregating LoRA parameters at the scale of the entire LoRA matrices, we perform aggregation at the decomposed rank-1 matrix level. For each rank-1 matrix, only the clients that update and upload that rank-1 matrix will participate in the aggregation process.
Each client \(k\) contributes to the global model based on a sparsity-based norm \(z_k\), defined as
\begin{equation}
z_k = 
\begin{cases} 
\|\mathbf{B}_{k,\text{dq}}^{\text{T}}\mathbf{A}_{k,\text{dq}}^{\text{T}}\|_F, & \text{Truncation scheme} \\
\|\mathbf{B}_{k,\text{dq}}^{\mathcal{I}}\mathbf{A}_{k,\text{dq}}^{\mathcal{I}}\|_F, & \text{Freezing scheme}
\end{cases},
\end{equation}
where \(\|\cdot\|_F\) denotes the Frobenius norm. The total norm list $Z$ has a length of $r_g$, and $Z[j]$ accumulates the $z_k$ values from all clients that have updated the $j$th rank-1 matrix. This list is then used to normalize $z_k$ to obtain the contribution weight for each rank-1 matrix.

\section{Numerical Results}
In our experiments, we utilize GPT2\cite{radford2019language} as the backbone model and test on the Banking77 classification dataset\cite{casanueva-etal-2020-efficient}. The experiment involves 10 clients, one base station, and one server, with the dataset evenly distributed among these 10 clients. We assume that different clients have diverse computational capabilities, which correspond to different LoRA ranks or freezing ratios.
Specifically, in the parameter truncation scheme, clients 0 to 2 have a LoRA rank of 2, clients 3 to 5 have a LoRA rank of 4, and clients 6 to 9 have a LoRA rank of 8. The global model at the server is configured with a LoRA rank of 8. 
In the parameter freezing scheme, all clients and the cloud server maintain a uniform LoRA rank of 8. However, clients 0 to 2 have a freezing ratio of 0.75, clients 3 to 5 have a freezing ratio of 0.5, and clients 6 to 9 have a freezing ratio of 0.
All 10 clients participate in the training during each communication round. Additionally, the clients are positioned at increasing distances from the base station, ranging from 1100 meters to 2000 meters in increments of 100 meters. 
Specifically, client 0 is located 1100 meters away, client 1 at 1200 meters, and so forth, with client 9 positioned at 2000 meters.
The hyper-parameters for the experiments are shown in Table \ref{tab:experiment-parameters}, while the wireless simulation parameters are detailed in Table \ref{tab:wireless-parameters}.
\begin{table}[h]
\centering
\renewcommand{\arraystretch}{1.2}
\caption{Experiment hyper-parameters.}
\label{tab:experiment-parameters}
\begin{tabular}{|l|l|l|l|}
\hline
\textbf{Parameter} & \textbf{Value} & \textbf{Parameter} & \textbf{Value} \\ \hline
LoRA Dropout & 0.1 & LoRA Bias & None \\ \hline
LoRA Scaling Factor & 32 & LoRA Initialization & Gaussian \\ \hline
LoRA Target Module & attn.c\_attn & Optimizer & Adam \\ \hline
Learning Rate $\eta$& 0.001 & Weight Decay $\lambda$ & 0.001 \\ \hline
Token Padding Length & 256 & Token Truncation & True \\ \hline
Smoothing $\beta_1$, $\beta_2$ & 0.85, 0.85 & Seed & 0, 1, 42 \\ \hline
\end{tabular}
\end{table}

\begin{table}[h]
\centering
\renewcommand{\arraystretch}{1.2}
\caption{Wireless system parameters.}
\label{tab:wireless-parameters}
\begin{tabular}{|l|l|l|l|}
\hline
\textbf{Parameter} & \textbf{Value} & \textbf{Parameter} & \textbf{Value} \\ \hline
Carrier Freq. $f$ & 2.4 GHz & Bandwidth $B_0$ & 100, 10 MHz \\ \hline
Commu. Time $T_{\text{ul}}$ & 10 ms & UE Tx Power $P_k$ & 23 dBm \\ \hline
PSD $N_0^{\text{bs}}$ & -174 dBm/Hz & Std. Dev. $\sigma$ & 7.8 dB \\ \hline
\end{tabular}
\end{table}

\begin{figure}[htbp]
\centering
\includegraphics[scale=0.38]{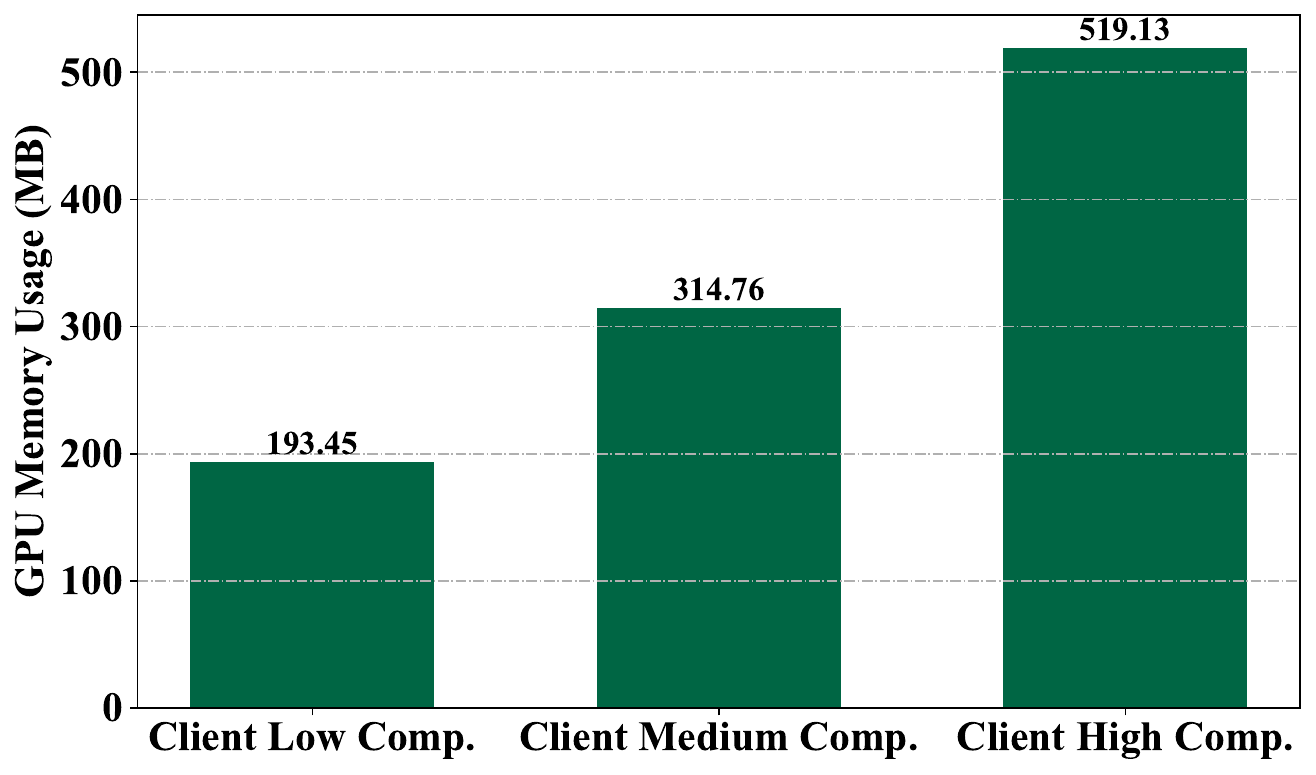}
\caption{GPU memory usage for clients with different computing capabilities.}
\label{gpt2_memory_usage}
\end{figure}

Fig.~\ref{gpt2_memory_usage} plots the GPU memory usage for clients with varying computational capabilities under our salience-driven adaptive LLM quantization algorithm. The GPT-2 model used in this experiment consists of 12 transformer blocks. For clients with low computational capabilities, all transformer blocks are quantized to 8 bits, resulting in the lowest GPU memory usage. For clients with medium computational capabilities, the algorithm ranks transformer blocks by salience, keeping 32-bit precision for the 6 most salient blocks while quantizing the remaining 6 blocks to 8 bits, achieving a balance between memory efficiency and model precision. For clients with high computational capabilities, all transformer blocks keep 32-bit precision, leading to the highest GPU memory usage. These results demonstrate the effectiveness of our proposed algorithm in dynamically adjusting the quantization strategy to optimize resource utilization while accommodating the varying computational capabilities of different clients.

\begin{figure}[htbp]
\centerline{\includegraphics[scale=0.49]{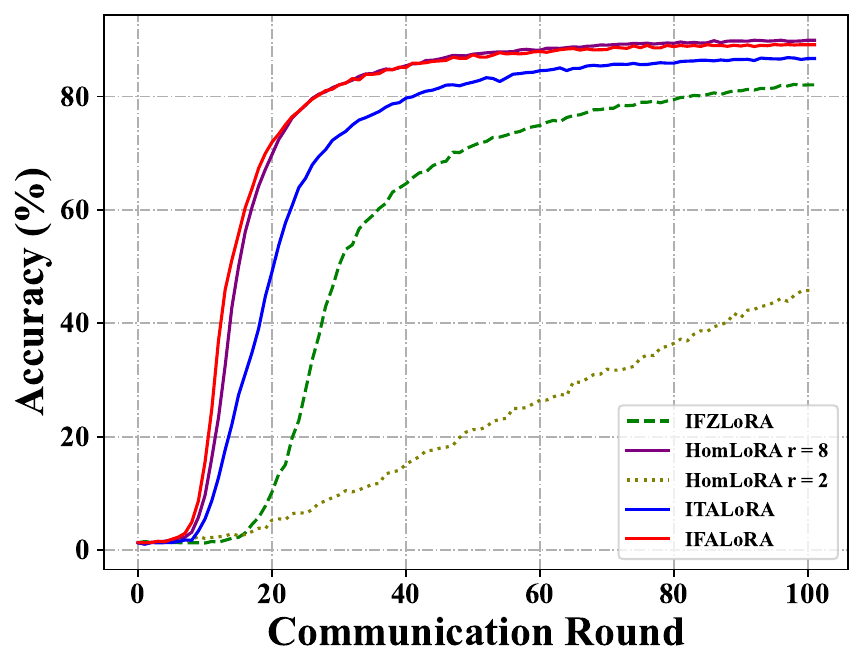}}
\caption{Accuracy comparison on Banking77 classification.}
\label{Accuracy_comparison}
\end{figure}

Fig.~\ref{Accuracy_comparison} presents the classification accuracy of the global model on the Banking77 classification test set as the number of communication rounds increases. In this experiment, we set the available bandwidth for each client to $B_0 = 100$ MHz, ensuring that communication bandwidth is not a limiting factor for FL parameter uploads. Our methods, importance-based parameter freezing adaptive aggregation LoRA (IFALoRA) and importance-based parameter truncation adaptive aggregation LoRA (ITALoRA), yield fast convergence. We also compared with three other algorithms: (1) homogeneous LoRA (HomLoRA) $r=2$, where all clients and the cloud server use LoRA with a rank of 2; (2) HomLoRA $r=8$, where all clients and the cloud server use LoRA with a rank of 8; and (3) IFZLoRA, which utilizes zero-padding aggregation compared to IFALoRA. The results indicate that HomLoRA $r=2$ fails to converge, and IFZLoRA converges slowly with significantly lower accuracy than our ITALoRA and IFALoRA. Our methods achieve accuracies close to that of HomLoRA $r=8$, where all clients use the highest rank.

\begin{table}[h]
\centering
\caption{Evaluation results on Banking77 classification.}
\label{tab:Evaluation_results_on_20_News_text_classification}
\begin{tabular}{l|l|c|cc}
\toprule
\multirow{2}{*}{\textbf{Model}} & \multirow{2}{*}{\textbf{Method}} & \multirow{2}{*}{\makecell{\textbf{Commu.} \\ \textbf{Size}}} & \multicolumn{2}{c}{\textbf{Classification Accuracy}} \\ \cline{4-5}
 &  &  & \textbf{50} & \textbf{100} \\ \midrule
\multirow{5}{*}{GPT2}
 & IFZLoRA & 7.03M & 71.34$_{\pm2.4}$ & 82.06$_{\pm1.2}$ \\ 
 & HomLoRA r=8 & 11.25M & 87.48$_{\pm0.0}$ & 89.94$_{\pm0.2}$ \\ 
 & HomLoRA r=2 & 2.81M & 21.26$_{\pm14.0}$ & 45.79$_{\pm16.0}$ \\ 
 & \textbf{ITALoRA} & 7.03M & 82.53$_{\pm2.1}$ & 86.70$_{\pm0.5}$ \\ 
 & \textbf{IFALoRA} & 7.03M & 87.36$_{\pm0.2}$ & 89.13$_{\pm0.2}$ \\ \bottomrule
\end{tabular}
\end{table}

Table \ref{tab:Evaluation_results_on_20_News_text_classification} presents the performance of five different methods on the Banking77 text classification task after 50 and 100 communication rounds. The values are averaged over experiments with different random seeds, and the table also records the average total parameter size uploaded by 10 clients per communication round. Specifically, HomLoRA with rank 8 achieves the highest accuracy but requires the largest communication size. In contrast, IFALoRA and ITALoRA manage to achieve nearly comparable accuracy with significantly smaller communication sizes, demonstrating a good balance between performance and efficiency. HomLoRA with rank 2 shows poor performance, indicating a lack of convergence, while IFZLoRA improves over HomLoRA r = 2 but still lags behind ITALoRA and IFALoRA in terms of accuracy.

\begin{figure}[]
    \centering
    \subfigure[]{
        \includegraphics[scale=0.46]{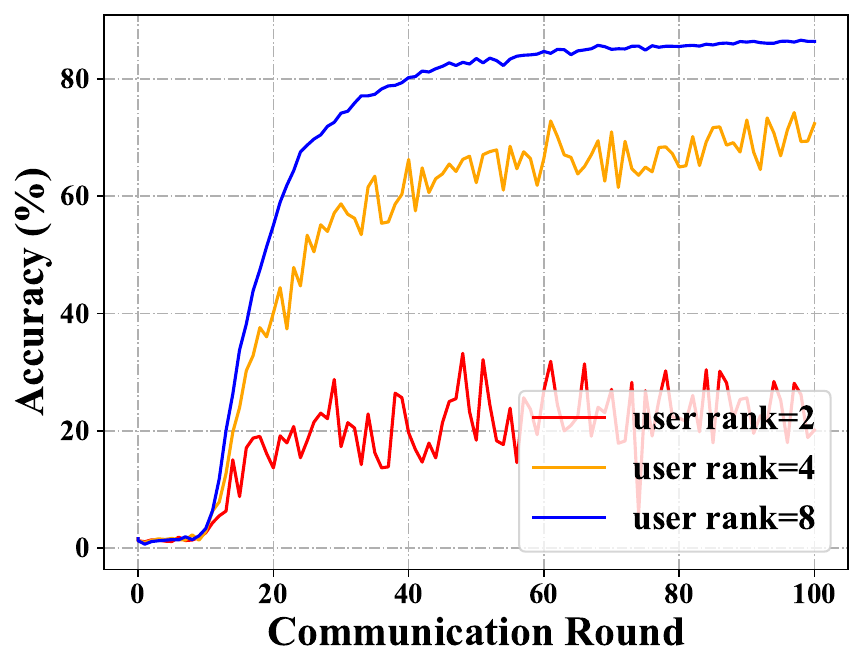}
        \label{fig:ITALoRA}
    }
    \subfigure[]{
        \includegraphics[scale=0.46]{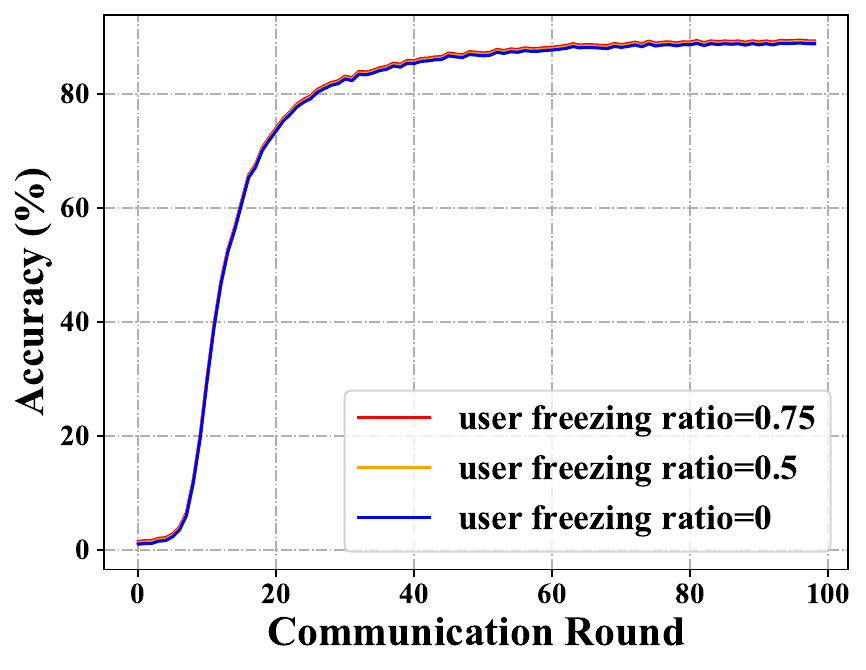}
        \label{fig:IFALoRA}
    }
    \caption{(a) Client local model performance in ITALoRA. (b) Client local model performance in IFALoRA.}
    \label{fig:Performance_degradation}
\end{figure}

Fig.~\ref{fig:Performance_degradation}~\subref{fig:ITALoRA} plots the classification accuracy with the increasing of communication rounds under ITALoRA when the global model is truncated to smaller client ranks.
As the level of truncation increases, the model's accuracy decreases, indicating a loss of information. In contrast, our IFALoRA method in Fig.~\ref{fig:Performance_degradation}~\subref{fig:IFALoRA} uses a freezing strategy, eliminating the need for truncation. This allows each client to maintain the same accuracy performance level as the global model. Additionally, both IFALoRA and ITALoRA have the same parameter size during training, ensuring communication efficiency while IFALoRA avoids the performance loss associated with truncation.

\begin{figure}[htbp]
\centerline{\includegraphics[scale=0.31]{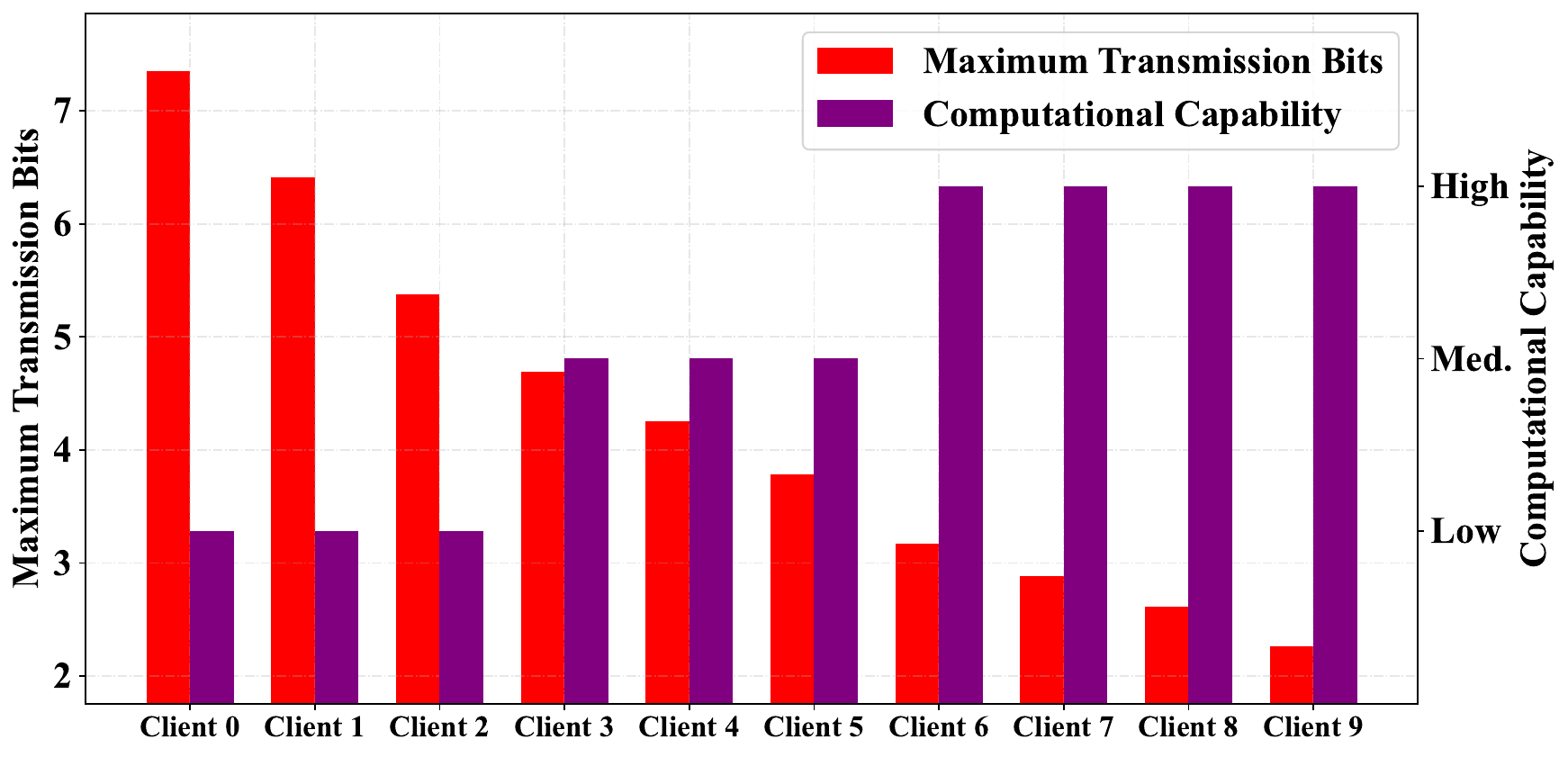}}
\caption{Maximum number of transmission bits and computational capability.}
\label{capacity_and_computing}
\end{figure}
Fig.~\ref{capacity_and_computing} depicts the distribution of the maximum number of transmission bits and computational capability among different clients. The red bars represent the maximum number of transmission bits, while the purple bars indicate the computational capability levels. With a fixed bandwidth allocation of $B_0=10$ MHz for each client, the maximum number of transmission bits vary based on the distance to the base station. Client 0, being the closest to the base station, can support the highest number of transmission bits. As the distance increases, the maximum number of transmission bits gradually decreases, with client 9 having the lowest value. For computational capability, we assign different levels to clients: clients 0 to 2 have the lowest computational capability, clients 3 to 5 are assigned a medium computational capability, and clients 6 to 9 have the highest computational capability.

\begin{figure}[htbp]
\centerline{\includegraphics[scale=0.33]{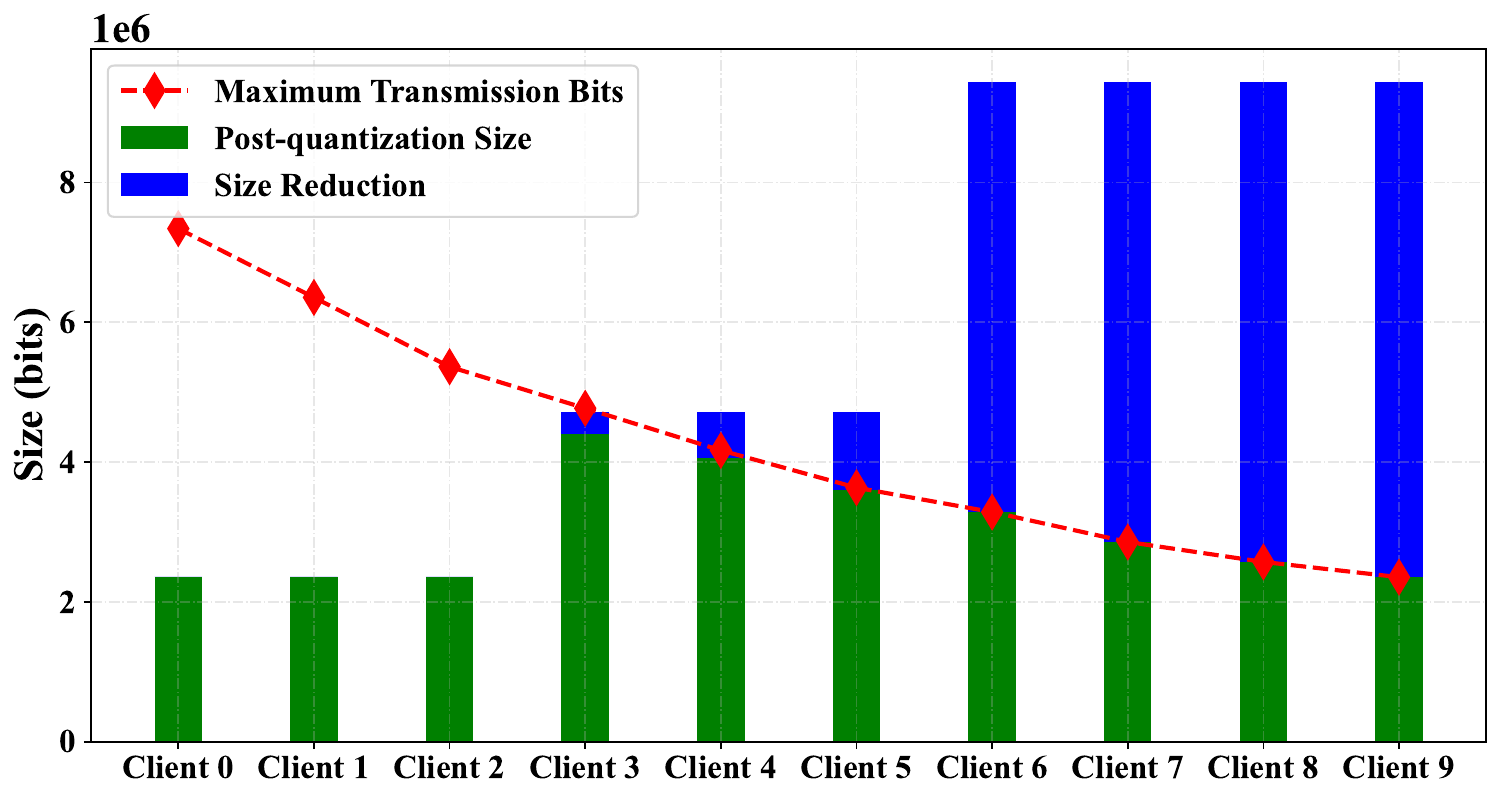}}
\caption{Importance-aware bandwidth-adaptive communication quantization scheme.}
\label{quantization_analysis}
\end{figure}
Fig.~\ref{quantization_analysis} plots the post-quantization communication size and the size reduction achieved for each client against the maximum transmission bits under the proposed importance-aware bandwidth-adaptive communication quantization scheme. 
Clients 0 to 2, which are relatively closer to the base station, benefit from higher maximum transmission bits and can upload their full-precision LoRA parameters without requiring quantization. In contrast, clients 3 to 9 have to be quantized to reduce the original LORA parameters' size under the limit of maximum transmission bits via our proposed adaptive quantization algorithm.

\begin{figure}[htbp]
\centerline{\includegraphics[scale=0.25]{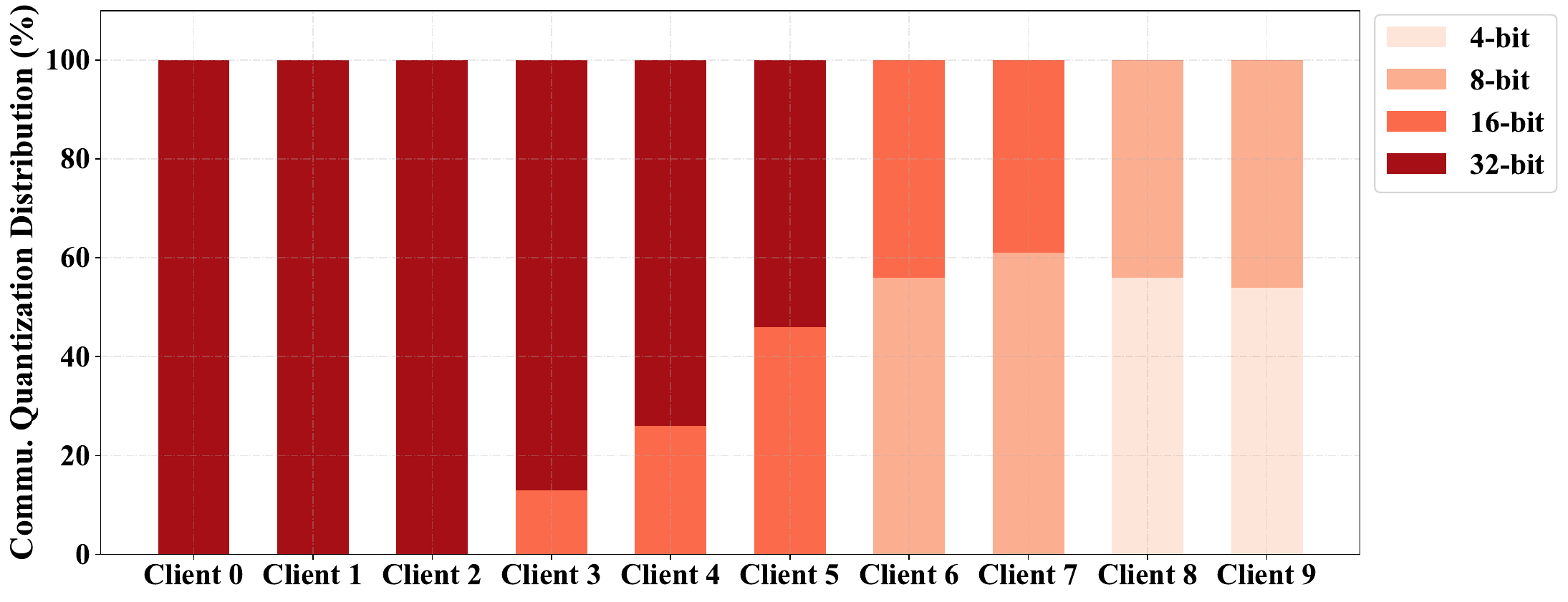}}
\caption{Communication Quantization distribution of importance-aware bandwidth-adaptive communication quantization.}
\label{precision_distribution}
\end{figure}

Fig.~\ref{precision_distribution} presents the communication quantization distribution across different clients under our importance-aware bandwidth-adaptive communication quantization algorithm. The quantization levels are dynamically adjusted based on the clients' communication conditions, leading to the following trends: Clients 0 to 2 are positioned close to the base station, benefiting from higher maximum transmission bits. As a result, they exclusively use 32-bit precision for transmission.
Clients 3 to 5 adopt a mixed-precision scheme combining 32-bit and 16-bit quantization. As the distance from the base station increases, the proportion of 16-bit precision gradually rises, reflecting the need for reduced communication overhead.
Clients 6 and 7 employ a mix of 16-bit and 8-bit quantization, further lowering communication costs while maintaining performance.
Clients 8 and 9 utilize the most aggressive quantization strategy, relying on a combination of 8-bit and 4-bit precision to stay within bandwidth constraints.

\begin{figure}[htbp]
\centerline{\includegraphics[scale=0.48]{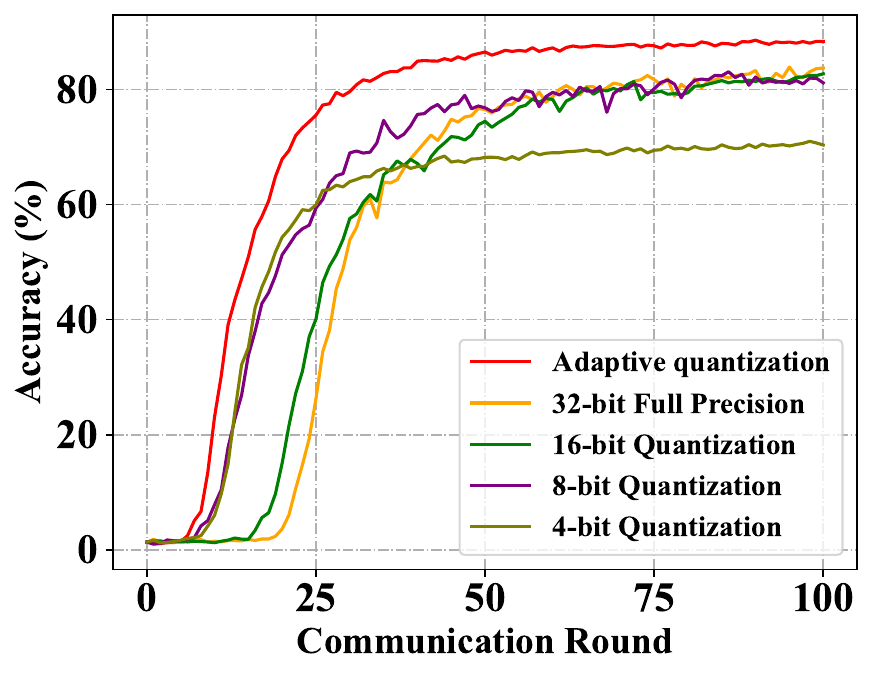}}
\caption{Accuracy comparison of different communication precision schemes.}
\label{Staged_quantization}
\end{figure}
Fig.~\ref{Staged_quantization} plots the classification accuracy comparison of different communication quantization schemes. All considered schemes are based on the IFALoRA framework, with different communication precision methods: full-precision (32-bit), half-precision (16-bit), 8-bit quantization, and 4-bit quantization. These methods maintain a fixed precision throughout training, prioritizing the transmission of important LoRA parameters first. If the number of parameters exceeds the bandwidth limit, the remaining ones are discarded.
Among all schemes, our proposed IFALoRA with importance-aware bandwidth-adaptive quantization scheme achieves the fastest convergence and the highest final accuracy. By dynamically adjusting the precision of the LoRA parameters based on bandwidth constraints, it ensures efficient communication while preserving model performance.
Among the fixed-precision quantization methods, the 4-bit quantization scheme achieves the fastest initial convergence due to its low communication overhead, which allows more parameters to be transmitted within the bandwidth limit. However, the high quantization error in 4-bit precision accumulates over communication rounds, leading to noisier updates and the lowest final accuracy. The 32-bit full-precision scheme, on the other hand, has the slowest convergence because of its high communication cost, which limits the number of parameter updates within the same budget. Nevertheless, it achieves relatively higher final accuracy by preserving all parameter information. The 16-bit and 8-bit quantization methods offer a good trade-off, achieving faster convergence than full precision while resulting in minimal loss in final accuracy.

\begin{figure}[htbp]
\centerline{\includegraphics[scale=0.31]{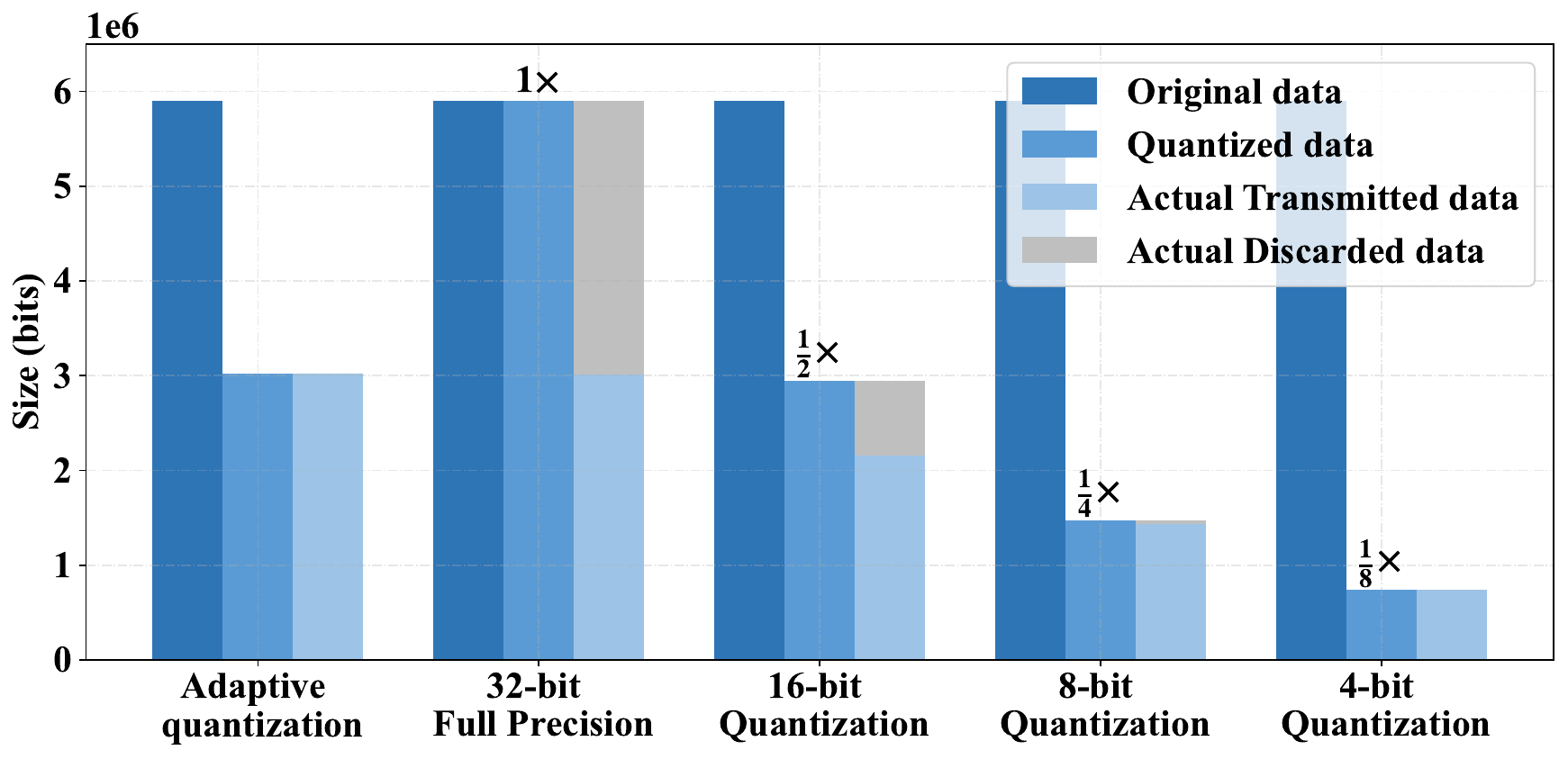}}
\caption{Average uplink data size of different communication precision schemes.}
\label{Staged_quantization_bar}
\end{figure}

Fig.~\ref{Staged_quantization_bar} presents a comparison of our importance-aware bandwidth adaptive communication quantization scheme against fixed-precision quantization methods in terms of the average uplink data size per client. 
The figure illustrates four key components: the original data size, the quantized data size, the actual transmitted data, and the discarded data due to bandwidth constraints.
Our adaptive quantization scheme ensures that the quantized data size precisely aligns with the available bandwidth, thereby eliminating any discarded data caused by bandwidth limitations. In contrast, the 32-bit full-precision scheme transmits significantly more data than the bandwidth can accommodate, resulting in nearly half of the data being discarded. As the bit width decreases, the proportion of discarded data gradually reduces, with the 16-bit and 8-bit quantization schemes exhibiting progressively lower levels of discarded data. The 4-bit quantization scheme remains entirely within the bandwidth limit, leading to no discarded data.
From a channel utilization perspective, both our adaptive scheme and the 32-bit full-precision scheme fully utilize the available bandwidth. However, while the 32-bit scheme results in substantial discarded data due to bandwidth constraints, our adaptive approach achieves optimal channel utilization by ensuring that all transmitted data fits within the available bandwidth, avoiding unnecessary data discard. In contrast, lower-precision schemes such as 4-bit quantization, despite preventing discarded data, do not fully exploit the available bandwidth, leading to reduced channel utilization.

\begin{figure}[H]
\centerline{\includegraphics[scale=0.28]{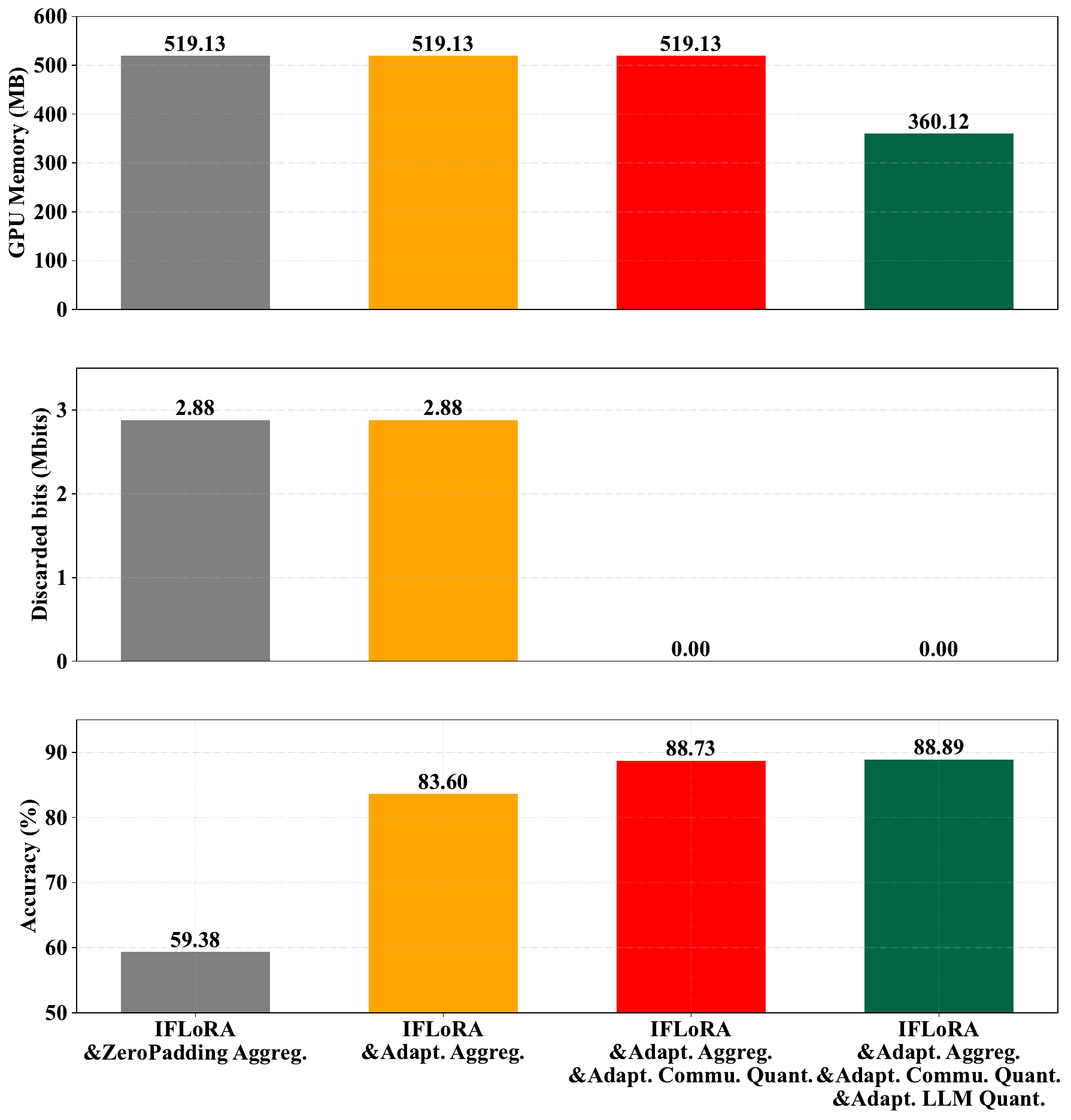}}
\caption{Comparison of performance metrics with incorporation of adaptive schemes.}
\label{all_adaptive}
\end{figure}

Fig.~\ref{all_adaptive} plots a comprehensive performance comparison when progressively incorporating adaptive aggregation, adaptive communication quantization, and adaptive LLM quantization, all built upon the foundation of the importance-based parameter freezing scheme, evaluated in terms of model accuracy, discarded communication bits, and GPU memory usage.
First, adaptive aggregation significantly improves accuracy from 59.38\% (under ZeroPadding aggregation) to 83.60\% by mitigating parameter dilution. Next, incorporating adaptive communication quantization further boosts accuracy to 88.73\% by dynamically adjusting quantization levels in response to bandwidth constraints, thereby eliminating discarded bits (reduced from 2.88 Mbits to 0). Finally, the inclusion of adaptive LLM quantization reduces GPU memory usage from 519.13 MB to 360.12 MB while maintaining high accuracy (88.89\%), thereby demonstrating its effectiveness in optimizing resource consumption without compromising performance.

\section{Conclusions}
In this paper, we proposed the Heterogeneous Adaptive Federated LoRA
Fine-tuned LLM with Quantization framework to address privacy concerns in fine-tuning LLMs. To accommodate client resource heterogeneity, we first designed a salience-driven adaptive quantization scheme that dynamically adjusts the proportion of quantized transformer blocks based on both the computational capabilities of the clients and the importance of the blocks.
To further address resource heterogeneity, we introduced an importance-based parameter truncation scheme at the local LLM. However, the truncation process inevitably resulted in performance degradation. To mitigate this, we developed an importance-based parameter freezing scheme at the local LLM. For the cloud server, we proposed an adaptive global model aggregation method to counter the information dilution problem caused by the zero-padding aggregation method.
Moreover, we proposed an Importance-Aware Bandwidth-Adaptive Communication Quantization scheme, which prioritized the transmission of important LoRA rank-1 matrices with higher precision under bandwidth constraints.
Experimental results demonstrated the effectiveness of our framework, with GPU memory usage decreasing by 31\%, communication cost reduced by 49\%, and model accuracy improving by 50\%. These results highlights that our framework effectively supports federated fine-tuning of LLMs in real-world heterogeneous environments.


\bibliographystyle{IEEEtran}
\bibliography{IEEEabrv,mylib}


\end{document}